\documentclass[11pt]{article}
\usepackage[T1]{fontenc}
\usepackage[utf8]{inputenc}
\usepackage{times}
\usepackage{latexsym}
\usepackage{amsmath,amssymb}
\usepackage{booktabs}
\usepackage{graphicx}
\PassOptionsToPackage{pdfusetitle}{hyperref}
\usepackage[preprint]{acl}

\newcommand{\PromptBioCross}{55.15}
\newcommand{\PromptBioWithinC}{26.53}
\newcommand{\PromptBioWithinS}{24.02}

\newcommand{\BioCommon}{34.97}
\newcommand{\BioScientific}{34.94}

\newcommand{\ErrorBioBase}{-6.32}
\newcommand{\ErrorBioConditional}{-2.47}

\newcommand{\ErrorBioRCI}{[-25.80, 17.45]}
\newcommand{\MarginBioLowShare}{91.38}
\newcommand{\MarginBioHighRate}{4.23}
\newcommand{\BioDisagree}{52.89}

\newcommand{\PromptHugeCross}{46.80}
\newcommand{\PromptHugeWithinC}{33.94}
\newcommand{\PromptHugeWithinS}{28.24}

\newcommand{\ErrorHugeBase}{-5.45}
\newcommand{\ErrorHugeConditional}{-1.51}

\newcommand{\ErrorHugeRCI}{[-25.72, 20.55]}
\newcommand{\MarginHugeLowShare}{75.79}
\newcommand{\MarginHugeHighRate}{4.08}

\newcommand{\DistractorBioFlip}{5.74}
\newcommand{\DistractorBioDelta}{-0.32}

\newcommand{\RareClasses}{395}
\newcommand{\RareImages}{11,833}
\newcommand{\RareGroups}{11,826}
\newcommand{\RareDevImages}{2,366}
\newcommand{\RareConfImages}{9,467}
\newcommand{\RareDevGroups}{2,364}
\newcommand{\RareConfGroups}{9,462}

\newcommand{\RareBioCross}{37.86}
\newcommand{\RareBioWithinC}{15.74}
\newcommand{\RareBioWithinS}{13.44}

\newcommand{\RareBioScientificAdvantage}{+12.11}

\newcommand{\RareEnsembleBioU}{36.13}
\newcommand{\RareEnsembleBioDeltaMatched}{-1.72}
\newcommand{\RareEnsembleBioDeltaDefault}{+0.47}
\newcommand{\RareHugeCross}{32.59}
\newcommand{\RareHugeWithinC}{21.41}
\newcommand{\RareHugeWithinS}{11.33}

\newcommand{\RareHugeScientificAdvantage}{+11.72}

\newcommand{\RareEnsembleHugeU}{28.57}
\newcommand{\RareEnsembleHugeDeltaMatched}{-4.03}
\newcommand{\RareEnsembleHugeDeltaDefault}{+1.65}

\newcommand{\RareBioPositiveExcessClasses}{363}
\newcommand{\RareHugePositiveExcessClasses}{339}
\newcommand{\URLClasses}{316}

\newcommand{\CachedReplayFiles}{181}
\newcommand{\CachedReplayIDs}{933,984}
\newcommand{\CachedReplayChecks}{5,826,648}
\newcommand{\URLIndependentChecks}{250,656}
\newcommand{\URLSgroupCrossDelta}{-5.53}
\newcommand{\URLSgroupWithinCDelta}{+5.03}

\newcommand{\URLOtherCrossDelta}{-5.65}
\newcommand{\URLOtherWithinCDelta}{+6.28}

\newcommand{\FreshAcquisitionFinding}{All retained frames pass the old-byte/RGB-overlap gate; the original failed check and pre-inference amendment remain archived.}

\newcommand{\FreshReproducibilityFinding}{An independent implementation reproduces all 2,370,384 fresh class decisions. A separate portable native run re-encodes all 5,487 images and 26,352 reviewed-pool text strings per checkpoint; the resulting fresh class decisions also have zero mismatches.}

\newcommand{\tablegraphic}[1]{\leavevmode\begingroup
  \setbox0=\hbox{#1}%
  \ifdim\wd0>\linewidth\resizebox{\linewidth}{!}{\box0}\else\box0\fi
  \endgroup}
\title{Beyond Naming Accuracy: Template Composition\texorpdfstring{\\}{ }and Species-Name Decisions}
\author{\texorpdfstring{%
  \normalfont Junle Li\textsuperscript{1,\ensuremath{\dagger},*},
  Bochang Zhang\textsuperscript{1,\ensuremath{\dagger}}%
}{Junle Li, Bochang Zhang}}
\date{}
\begin{document}
\maketitle
\footnotetext[1]{The University of Edinburgh.\newline
  \textsuperscript{\ensuremath{\dagger}}Equal contribution and shared first authorship.\newline
  \textsuperscript{*}Corresponding author
  \href{mailto:junle8344@gmail.com}{\texttt{junle8344@gmail.com}}}
\setcounter{footnote}{1}
\begin{abstract}
Vision-language classifiers identify species by matching images to text prompts containing species names. BioCLIP2 changes its predicted species less often than BioCLIP 2.5 Huge across four photographic templates, but the ranking reverses across a bank of 80 templates. This comparison leaves open whether the reversal depends on the number of templates or their wording. We hold the count at four and compare the classifiers on every four-template subset of the 80-template bank and of a second pool of 24 reviewed photographic templates, keeping images and candidate names fixed. Huge is more consistent across the reviewed pool as a whole, yet 19.09\% of its four-template subsets favor BioCLIP2 on iWildCam and 5.10\% do so on Rare Species. We then test whether both rankings hold on 5,487 frames from previously unused iWildCam sequences. Among 4,356 subsets containing two \emph{photo} and two \emph{photograph} templates, simultaneous 95\% intervals support lower disagreement for Huge in 1,526 subsets and for BioCLIP2 in 43. Even with the same prompt count and noun mixture, the choice of sentences can determine which classifier appears more consistent.
\end{abstract}

\section{Introduction}
Vision-language models identify species by matching images to prompts that contain candidate names \citep{gu2025}. The surrounding sentence can change the predicted species and may affect which model appears more consistent. In this work, we compare BioCLIP2 and BioCLIP 2.5 Huge across every four-template subset of two prompt pools, holding images and candidate names fixed. We then test for a ranking reversal on frames from previously unused camera-trap sequences. For two sets that each contain two \emph{photo} and two \emph{photograph} templates, Huge's disagreement is 16.38 percentage points lower than BioCLIP2's under one set and 8.31 points higher under the other.

Why examine changes in the predicted species? In the original iWildCam audit, BioCLIP2 attains almost identical accuracies with common and scientific names in its default sentence, 34.97\% and 34.94\%. Yet the predicted category changes on 52.89\% of frames under the audit weights. The common name alone matches the source label on 7.22\% and the scientific name alone on 7.19\%. These changes cancel almost exactly in the accuracy difference. On another 38.48\%, the two predictions name different categories and both miss the label. A comparison of accuracies loses precisely the information needed to answer whether a reader would see the same species name twice.

This distinction raises a second question. To call one checkpoint more consistent than another, which changes of wording should count? For common names, an audit with four ordinary photographic sentences favors BioCLIP2, while an audit with the complete 80-template ImageNet bank favors Huge. The larger bank changes both the number of sentences and the kinds of depictions they describe, including toys and paintings. To distinguish these explanations, we use a \emph{pool} to denote the declared set of eligible templates and a \emph{menu} to denote the subset used for a particular comparison. A four-template menu gives six pairs of predictions to compare on each image.

Our main contributions are as follows:
\begin{enumerate}
\item \textbf{Holding count fixed preserves the bank reversal.} Among all four-template menus from the official bank, 96.73\% favor Huge on iWildCam and 96.14\% on Rare Species. Four sentences suffice to recover the direction seen in the full bank.
\item \textbf{Restricting the wording still leaves both orderings.} In the reviewed photographic pool, 722 of 4,356 equally noun-balanced menus favor BioCLIP2 on iWildCam, as do 227 on Rare Species, even though the complete pool favors Huge on both.
\item \textbf{Both orderings survive a change of images.} On previously unused iWildCam sequences, intervals covering all 10,626 reviewed four-template menus simultaneously support opposite directions within the balanced subset. Of its 4,356 menus, 43 favor BioCLIP2, 1,526 favor Huge, and 2,787 have unresolved directions.
\end{enumerate}
The first two findings describe the observed images and motivate the third, which uses a protocol fixed before fresh predictions. A separate evaluation-budget study then asks how much evidence is lost when fewer frames per class are available.

\section{Related work}
\paragraph{Species names as classifier inputs.}
CLIP constructs category prototypes from text \citep{radford2021}. In biological recognition, the text can be a common name, a scientific name or a taxonomic description. \citet{parashar2023} compare the recognition accuracy of common and scientific names across sources. TAILS examines synonyms and neglected concepts \citep{parashar2024tails}, and IchthyoNoma studies naming, prompts and visual context for fish recognition \citep{ichthyonoma2026}. BioCLIP and BioCLIP2 bring biological imagery and taxonomic supervision to this setting \citep{stevens2024,gu2025}. We compare BioCLIP2 with the later BioCLIP 2.5 Huge checkpoint \citep{bioclip25huge2026}. Keeping the candidate names fixed lets us examine how often the surrounding sentence alone changes the winning category.

\paragraph{Prompt choice as part of evaluation.}
\citet{kostiuk2026one} show that random and favorable prompt choices can change embedding-model rankings across six models and eleven datasets. PTEB uses stochastic paraphrases and human validation to evaluate embedding models under wording changes \citep{frank2026pteb}. For a classifier that returns a species ID, each prompt produces a discrete decision that we can pair with a second decision on the same image. This gives us two explicit sampling questions. How much would the checkpoint contrast change with another menu from the same pool? With a particular menu fixed, how much would it change with another sample of images? We address the first by enumeration and the second by stratified sampling.

\section{Comparing paired decisions}
\label{sec:v10-measurements}
\paragraph{From one prediction to a menu.}
Let $p_i^m(a,t)$ be the category predicted on image $i$ by checkpoint $m$, using name type $a$ and template $t$. To compare two sentences that contain the same common names, we record
\[
 u_i^m(a;t,t')=\mathbf 1[p_i^m(a,t)\ne p_i^m(a,t')].
\]
Averaging over the $\binom{k}{2}$ unordered pairs of a $k$-template menu $B$ and then over images gives its disagreement rate $U^m_a(B)$. For example, three identical decisions and one different decision on a frame produce three disagreeing pairs out of six. A two--two split produces four. Both patterns break unanimity, but the second gives a reader a higher chance of receiving different names from two distinct templates chosen uniformly.

We compare checkpoints by subtracting BioCLIP2's disagreement from Huge's, $D_a(B)=U_a^{\rm Huge}(B)-U_a^{\rm Bio2}(B)$. A positive difference favors BioCLIP2's consistency. All differences in rates are in percentage points (pp). Each checkpoint sees the same pairs and receives the same image weights. The primary analysis holds the common names fixed ($a=C$). For the separate naming comparison, we hold the sentence fixed, substitute scientific for common names across the vocabulary, and average the resulting paired disagreements over sentences. Accuracy is the fraction of predictions matching the source label under the same image weights.

A change of template is applied to every candidate, so it can change the winner even when the score of the image's labelled species hardly moves. An earlier diagnostic illustrates the role of competitors. Replacing only the common name \emph{gray four-eyed opossum} with its scientific counterpart for every image changed \DistractorBioFlip\% of BioCLIP2 decisions on iWildCam, although none of the target frames carried that source label (Appendix~\ref{app:replacement}). This is why we keep the full candidate vocabulary throughout the menu comparisons.

\paragraph{From a menu to a pool.}
A comparison based on one menu gives each of its pairs equal weight and all other pairs zero weight. If $B$ is drawn uniformly from the $k$-subsets of an $n$-template pool $T$, each pair is included with probability $\binom{k}{2}/\binom{n}{2}$. It follows that
\begin{equation}
 \mathbb E_B[D_a(B)]=D_a(T).
 \label{eq:menu-mean}
\end{equation}
Changing $k$ leaves this expectation fixed. It can still change the spread across individual menus. We compute the image-weighted contrast for each pair once, then average the six corresponding entries for every four-template menu. This gives an exhaustive distribution while retaining pairs that share a template or an image. Appendix~\ref{app:v10-menu-moments} gives the exact variance at other budgets. The identity applies to the disagreement statistic. Averaging text prototypes before taking a single winning category defines a different prediction rule.

\paragraph{From the observed images to a population.}
Enumeration accounts for every menu on the images we scored. Extending a menu's sign to the full camera-trap inventory requires an additional calculation. In the prospective experiment, let $C=27$ be the number of supported classes and $m_c$ the number of frames drawn without replacement from the $N_c$ eligible frames of class $c$. We average within class and give each class weight $1/C$. For a family of $M$ contrasts whose frame-level values have range width $R$, a conservative simultaneous 95\% half-width is \citep{hoeffding1963}
\begin{equation}
 h=R\sqrt{\frac{\log(2M/.05)}{2}
           \sum_{c:m_c<N_c}\frac{1}{C^2m_c}}.
 \label{eq:fresh-radius}
\end{equation}
A fully observed class contributes to the estimate and has zero sampling uncertainty. For example, the new inventory contains five frames for class 99 and 11,009 for class 372. We observe all five of the former and 256 of the latter. Their class means carry equal weight, but only the latter enters the sum in Equation~\ref{eq:fresh-radius}. The calculation relies on independent draws across strata and conditions on their finite sets of frame responses, allowing frames from the same camera burst to look alike.

\section{Experiments}
\label{sec:v10-design}
Our evaluation is designed to determine whether a checkpoint ranking survives progressively tighter controls on prompt choice. We therefore begin on two image sources, iWildCam2022 and Rare Species, and reserve previously unused iWildCam sequences for confirmation.

\noindent\textbf{RQ1.} Can the prompt pool change the ranking at a fixed template count? (Section~\ref{sec:rq1})

\noindent\textbf{RQ2.} Can the ranking reverse with the same mix of photographic nouns? (Section~\ref{sec:rq2})

\noindent\textbf{RQ3.} Are both rankings supported on new camera-trap sequences? (Section~\ref{sec:rq3})

\noindent\textbf{RQ4.} How does the frame budget affect support for both rankings? (Section~\ref{sec:rq4})

\paragraph{Implementation details.}
We run both released checkpoints in FP32 with their native tokenizers and official image preprocessing, including the center crop. Within each checkpoint, all prompts share a frame's image features. We normalize image and text features and take the candidate with largest cosine similarity, breaking exact ties by ascending source category ID. All 154 iWildCam or 395 Rare candidate IDs compete for every frame from their respective source, including classes outside the image-sampling target. A separate original audit includes a general CLIP reference. The study compares released systems with different architectures, training data and preprocessing. Model-specific and source-specific details for the original audits are in Appendix~\ref{sec:design}.

\subsection{RQ1. Can the prompt pool change the ranking at a fixed template count?}
\label{sec:rq1}
\begin{figure*}[t]\centering
\includegraphics[width=\textwidth]{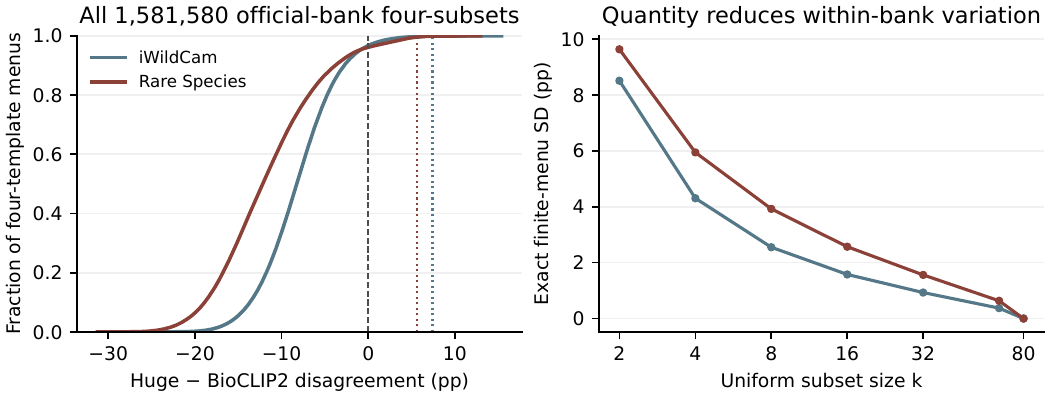}
\caption{Official-bank contrasts on existing images. Left, cumulative distribution of all four-template menus with the original four shown by dotted lines. Right, exact standard deviation across uniform subsets of each size.}
\label{fig:quantity}
\end{figure*}

The original iWildCam confirmation sample consists of 6,740 frames drawn from 81,969 in 28 supported classes \citep{iwildcam2022}. Within each class we restore the population proportions of the two detector strata, then average classes equally. Rare Species contributes 9,467 confirmation images from 395 classes following exclusions for ambiguous source names and a split that keeps exact duplicate groups together \citep{rarespecies2023}. We average over the entire fixed Rare cohort with equal class weights. Its rates are descriptive. Images, candidates and weights are held fixed when switching templates within each source.

We begin with the four sentences used in the original audit. They frame the name as \emph{a photo of a}, \emph{a photograph of a}, \emph{a photo showing a}, or \emph{this is a photo of a}, followed by the name and a period. Huge has lower common/scientific disagreement on average across these sentences. When the common names are held fixed and the sentence changes, Huge has \emph{higher} disagreement than BioCLIP2 by 7.41 pp on iWildCam and 5.66 pp on Rare. Moving to the complete 80-template ImageNet bank from the official demo \citep{bioclip2demotemplates2026} reverses that comparison to $-8.21$ and $-11.70$ pp. The larger bank varies depiction medium and visual conditions as well as phrasing.

The type of language change matters already at this baseline. In the original four sentences, matched common/scientific disagreement on iWildCam averages 55.15\% for BioCLIP2 and 46.80\% for Huge. Yet in the sentence with \emph{a photograph of a}, the respective rates are 57.44\% and 58.62\% (Table~\ref{tab:prompt}). One sentence can thus give an ordering opposite to the four-sentence mean even for a name substitution. For the rest of RQ1 we keep the names fixed, so changing the surrounding sentences is the only change to the text interface.

\paragraph{Equal-sized comparisons.}
Of the $\binom{80}{4}=1{,}581{,}580$ official-bank menus, 96.73\% on iWildCam and 96.14\% on Rare already give a negative contrast. Figure~\ref{fig:quantity} places the original four well into the positive tail on both sources. Only the default sentence occurs verbatim in both pools, so the dotted lines are external reference values. Removing that one sentence from the official bank preserves its negative mean contrast on both sources (Appendix~\ref{app:external79}).

\paragraph{What a larger prompt budget buys.}
Increasing the number of official templates from four to 16 reduces the exact across-menu standard deviation from 4.30 to 1.58 pp on iWildCam and from 5.94 to 2.57 pp on Rare. At 80, the menu is fully specified and this variation vanishes. Equation~\ref{eq:menu-mean} explains why the center stays fixed throughout. These results answer RQ1 and expose a limit of the comparison. They show that count alone cannot account for the reversal. A remaining possibility is that the result depends on the descriptions of non-photographic media in the official bank.

A literal word filter offers only a partial control. In an exploratory partition of the official bank, the 43 sentences containing the word \emph{photo} still give a negative mean within-common contrast when paired with each other on both sources (Appendix~\ref{app:photo-word}). A photograph of a toy can pass that filter while changing what the photograph depicts. To ask whether different rankings are possible under ordinary photographic descriptions, we therefore specify the intended meaning before scoring the next pool.

\subsection{RQ2. Can the ranking reverse with the same mix of photographic nouns?}
\label{sec:rq2}
We construct a pool in which every sentence describes a photograph of the named category. Table~\ref{tab:v10-wording} lists the 12 sentence forms. We instantiate each with both photographic nouns, giving 24 templates. Unlike the original four, these forms omit the article immediately before the name.

\begin{table}[!htbp]\centering
\tablegraphic{\includegraphics{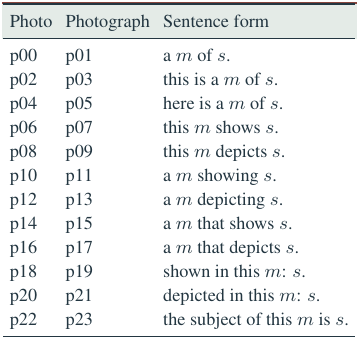}}
\caption{Reviewed templates. $m$ is the photographic noun and $s$ is the species name.}
\label{tab:v10-wording}
\end{table}

Across the reviewed pool, Huge has lower within-common disagreement on both sources, 37.25\% versus 40.76\% for BioCLIP2 on iWildCam, and 23.72\% versus 29.54\% on Rare. But the 24-template average combines 276 pairs, only six of which would enter a four-template evaluation. Among all 10,626 four-template menus, 2,029 on iWildCam (19.09\%) and 542 on Rare (5.10\%) have a positive contrast. Thus a single four-sentence comparison can disagree with the full-pool result even when every sentence describes a photograph.

The reviewed pool also separates two ways to perturb a sentence. With the noun fixed there are 66 pairs of distinct sentence forms for each noun. Across nouns there are 144 pairs, of which only 12 compare the same form. The full-pool average therefore gives substantial weight to pairs that change both the noun and the syntax. Reading the average as the effect of replacing one word would assign those other changes to the noun.

A possible explanation for the positive four-menu contrasts is that these menus contain different proportions of \emph{photo} and \emph{photograph}. To control that mixture, we retain all 4,356 menus with two of each. Every one now contains a photo--photo pair, a photograph--photograph pair and four cross-noun pairs. Sentence forms can still differ. On iWildCam, 722 of these menus (16.57\%) favor BioCLIP2 and the estimated contrasts range from $-16.74$ to $+9.20$ pp. On Rare, 227 (5.21\%) favor BioCLIP2, with a range of $-13.74$ to $+5.63$ pp. The noun balance therefore leaves room for either point ordering. Since we inspected these images to formulate that observation, we next test whether a new frame draw supports it.

\subsection{RQ3. Are both rankings supported on new camera-trap sequences?}
\label{sec:rq3}
We exclude every sequence represented among the 7,625 previously evaluated iWildCam frames. The resulting inventory has support for 27 of the original 28 target classes. From each supported class we draw up to 256 frames without replacement. Before inference, an identity check detects two old-image duplicates in the 5,489-frame draw. We apply a documented CRC/size screen to the entire residual inventory, removing six entries including those two. Intersecting the original draw with that screened domain yields 5,487 frames from 3,313 sequences in a target of 52,145 frames. We keep all 154 candidate names. Conditional on the retained counts, the frames are a simple random sample within each screened class. Appendix~\ref{app:v10-protocol} records the failed identity check, amendment and class sizes.

\paragraph{Do the bank-level results transfer?}
Before the new predictions, we fix three bank comparisons and four photographic-wording comparisons as separate 95\% families. Table~\ref{tab:v10-fresh} gives all seven estimates. For the original four sentences, disagreement is 25.37\% for BioCLIP2 and 32.57\% for Huge. In the official bank the corresponding rates rise to 58.17\% and 49.79\%. Both checkpoints change category more often across the broader bank, but the larger increase for BioCLIP2 reverses their order. The paired change in the checkpoint contrast is $-15.58$ pp, with interval $[-21.66,-9.50]$. This result concerns the screened 27-class population.

\begin{table}[!htbp]\centering
\tablegraphic{\includegraphics{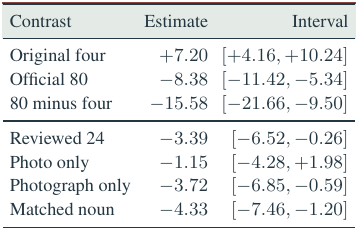}}
\caption{New-sequence disagreement contrasts (Huge minus BioCLIP2, pp). Intervals are simultaneous within each block.}
\label{tab:v10-fresh}
\end{table}

Taking the bank difference on each frame preserves the pairing between the two evaluations. A frame on which the two checkpoints are equally sensitive under both banks contributes zero to the bank change, however often either checkpoint switches names. Each bank's checkpoint contrast lies between $-1$ and $+1$ per frame. Subtracting the two contrasts gives a range from $-2$ to $+2$, which explains why the third interval in the table is wider than the first two. All three bank estimates have support in their anticipated directions.

The lower block distinguishes changes of sentence form from changes of noun. Across all 24 templates, Huge has lower disagreement by 3.39 pp. The photograph-only interval also lies below zero, while the photo-only interval includes zero. These two intervals do not directly test a difference between the nouns. A cleaner substitution holds the sentence form fixed and compares its two noun variants. Across the 12 such pairs, the contrast is $-4.33$ pp, with interval $[-7.46,-1.20]$. Huge changes its decision less often on average under that specific substitution.

The pool-wide ordering also need not hold for each species. In a descriptive check of the new frames, 23 of 27 class contrasts are positive under the original four and 22 are negative under the official bank. In the reviewed pool, 19 are negative and eight positive. Dropping one class at a time keeps the reviewed-pool mean between $-4.08$ and $-1.98$ pp. This check varies the class included in the average. The next test keeps all 27 classes and their weights fixed while varying only the four sentences.

\paragraph{Can equally balanced menus support opposite rankings?}
After the existing-image analysis and before new image encoding or decision scoring, we fixed a third family comprising all 10,626 reviewed four-menus. Its success criterion required at least one balanced menu with a lower bound above zero and another with an upper bound below zero. All menus use the same 5.00 pp simultaneous half-width from Equation~\ref{eq:fresh-radius}. This permits reporting the most positive and most negative balanced menus after enumeration while accounting for their selection. The three families give separate coverage guarantees.

\begin{table}[!htbp]\centering
\tablegraphic{\includegraphics{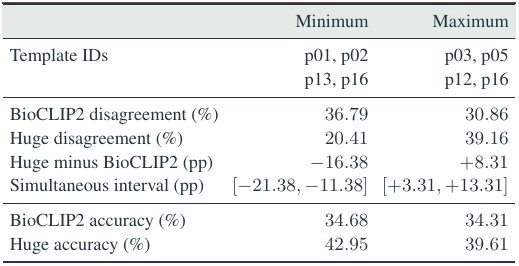}}
\caption{Balanced menus with the lowest and highest new-sequence contrasts. Template IDs refer to Table~\ref{tab:v10-wording}.}
\label{tab:v10-witnesses}
\end{table}

\begin{figure}[!htbp]\centering
\includegraphics[width=\linewidth]{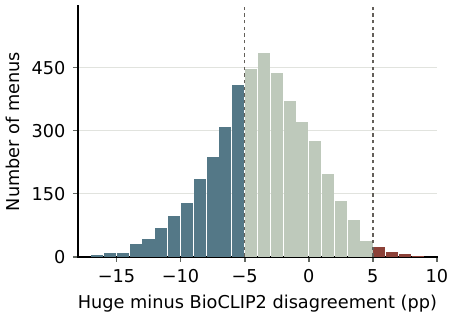}
\caption{Contrasts for 4,356 balanced menus on new frames. Dashed lines mark the simultaneous 5.00 pp radius on either side of zero.}
\label{fig:balanced-distribution}
\end{figure}

Figure~\ref{fig:balanced-distribution} shows where the supported menus lie. The center contains both positive and negative estimates whose intervals reach zero. The tail to the left of the negative threshold is considerably larger than the one to the right of the positive threshold. Of 4,356 balanced menus, 771 have a positive point estimate. Only 43 have intervals entirely above zero. Another 1,526 have intervals entirely below zero, leaving 2,787 unresolved. The criterion for both directions is therefore met. Table~\ref{tab:v10-witnesses} shows what changes at the extremes. Moving from the minimum to the maximum menu lowers BioCLIP2's disagreement from 36.79\% to 30.86\% and raises Huge's from 20.41\% to 39.16\%. One sentence, \emph{a photo that depicts} the named category, appears in both menus. Since the other three sentences change together, their individual contributions cannot be read from the difference between these two menus.

\subsection{RQ4. How does the frame budget affect support for both rankings?}
\label{sec:rq4}
RQ1 varied the template budget while holding the images fixed. We now hold the template pool, both checkpoints and the two-direction criterion fixed and vary the number of scored frames. This is a retrospective sensitivity study on the completed new-sequence cohort. Within each class, we randomly permute the retained frames and take nested prefixes capped at 16, 32, 64 and 128 frames. Shorter classes contribute all available frames. Each of 200 permutations yields one draw per budget, shared by every menu and both checkpoints. For each draw we recompute the class-weighted pair contrasts and compare all balanced menus with the simultaneous radius for the full 10,626-menu family, using the reduced sample counts and the original screened population sizes. The full draw provides a deterministic reference.

\begin{table}[!htbp]\centering
\tablegraphic{\includegraphics{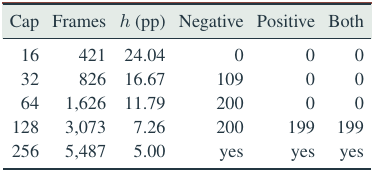}}
\caption{Frame-budget sensitivity. $h$ is the simultaneous radius. Counts are out of 200 subsamples, except for the single full draw.}
\label{tab:frame-budget}
\end{table}

At 64 frames per class, 1,626 frames are scored and the radius is 11.79 pp. All 200 draws support a negative contrast for at least one balanced menu, yet none supports a positive contrast. At the smaller cap of 32 frames, even the negative direction is supported in only 109 draws. Thus the loss of the two-direction result has a specific source at the intermediate budget: the narrower positive tail is harder to resolve. At 128 per class, 3,073 frames are scored, the radius falls to 7.26 pp, and 199 draws support both directions. Using the full 5,487 frames reduces the radius further to 5.00 pp (Table~\ref{tab:frame-budget}). The contrast that favors BioCLIP2 in the strongest full-sample example is only $+8.31$ pp, which helps explain why the positive direction is difficult to resolve with a wide interval. The reduced draws also change the point estimates, so we recompute them at each cap. These frequencies describe subsamples of this cohort. They do not supply a minimum sample size for new camera sites or independent replications of RQ3.

The two budgets in the study serve different purposes. Scoring more templates on the same frames gives a more stable average over a specified pool, as the standard deviations in Figure~\ref{fig:quantity} show. Scoring more frames for fixed menus can distinguish estimates that previously fell inside the sampling radius. For example, retaining all 24 templates at the 64-frame cap allows us to enumerate every menu exactly on the subsample, but that complete enumeration still fails to resolve the positive direction in any of the 200 draws. An evaluator deciding where to spend an additional batch of encodings needs to know which source of variation limits the comparison.

\section{Discussion}
\paragraph{Choosing a checkpoint for a stated objective.}
The same balanced menus illustrate why a consistency audit should retain accuracy alongside disagreement. In the minimum menu of Table~\ref{tab:v10-witnesses}, Huge is both more consistent and more accurate against the source labels (42.95\% versus 34.68\%). In the maximum menu it still has higher descriptive accuracy (39.61\% versus 34.31\%), but BioCLIP2 is more consistent. The decision changes for an evaluator who prioritizes repeatability across phrasings, even though the accuracy preference is the same. The simultaneous bounds cover disagreement. Choosing a deployment objective would additionally require a justified cost for a wrong species label or a changed prediction.

The appropriate summary also depends on how prompts will be used. For an interface that always offers the same four sentences, the six pair comparisons for those sentences answer the user's question directly. For an interface that offers all 24 reviewed sentences with equal probability and compares distinct wordings, the full-pool pair average answers a different question. Reporting only the best or worst four-menu result for the latter interface would give every other available sentence zero weight. The menu distribution shows how much is lost by making that substitution.

\paragraph{What changes if prompts are ensembled?}
A classifier can average four normalized text vectors for each candidate and issue one prediction per frame. That uses the four prompts for prediction, whereas our audit uses their separate predictions to measure disagreement. In the original retrospective policy comparison, development accuracy selected a single template in all six checkpoint--source combinations when singles and ensembles were both eligible. On confirmation, the selected BioCLIP2 ensemble scored 1.19 pp below the selected single on iWildCam and 0.31 pp above it on Rare (Table~\ref{tab:selection}). Ensemble accuracy and pairwise consistency thus address different decisions an evaluator might face with the same prompt budget.

\paragraph{What changes if images are filtered?}
Template choice determines which language changes enter an average. Frame selection determines which images receive weight. For the original common/scientific default on iWildCam, detector selection reduced the proportion of frames receiving two different wrong BioCLIP2 labels by 8.79 pp. That rate depends both on how often both policies miss and on how often their wrong predictions differ given a double miss. For the latter conditional rate the change was $-4.67$ pp with an interval of $[-25.80,+17.45]$ (Appendix~\ref{app:inference}). Reporting the image population alongside the prompt pool makes clear which of these statements the data support.

\section{Conclusion}
Which checkpoint returns a more consistent species name depends on the sentences an evaluator considers interchangeable. On both existing image sources, matching the prompt budget leaves a gap between the original photographic sentences and the broader official bank. Narrowing the comparison to reviewed photographic wording and matching the nouns still leaves menus favoring either checkpoint. On new sequences, simultaneous bounds support both directions. Reporting the pool, the distribution over equal-sized menus and the uncertainty from sampled images makes the scope of a checkpoint comparison explicit.

\section*{Limitations}
Source labels define both the accuracy calculation and the classes used for weighting. There was no expert verification of visible organisms. Rare includes specimen and catalogue material, and the native image crop may remove visual evidence. The checkpoint comparison consequently describes decisions on the supplied inputs. Differences in training, architecture and preprocessing prevent assigning the reversal to a specific component.

The fresh frames come from previously unused sequences in one screened iWildCam inventory. Camera sites may overlap earlier data and pretraining overlap is unknown. The population amendment followed a failed identity check and preceded inference. Each nominal 95\% family covers its declared comparisons on that finite population. The budget study reuses that same draw and cannot establish how often the result would recur in a newly collected survey.

\section*{Ethical considerations}
We use released benchmark data and source metadata. The retained qualitative illustrations preserve their publisher-provided license and credit fields. We do not redistribute the raw photo collection, model weights, exact GPS coordinates or new animal-location inferences. Source-label outputs are not validated species-identification or conservation advice.

\bibliography{references}
\clearpage
\appendix
\raggedbottom
\setcounter{topnumber}{6}
\setcounter{totalnumber}{10}
\setcounter{dbltopnumber}{10}
\renewcommand{\topfraction}{.95}
\renewcommand{\textfraction}{.05}
\renewcommand{\floatpagefraction}{.8}
\renewcommand{\dbltopfraction}{.95}
\renewcommand{\dblfloatpagefraction}{.8}

\makeatletter
\setlength{\@fptop}{0pt}
\setlength{\@fpsep}{14pt}
\setlength{\@fpbot}{0pt plus 1fil}
\setlength{\@dblfptop}{0pt}
\setlength{\@dblfpsep}{14pt}
\setlength{\@dblfpbot}{0pt plus 1fil}
\makeatother

\section{Original image populations and prediction policies}
\label{sec:design}
\subsection{Source references and image units}
The iWildCam2022 training release \citep{iwildcam2022} supplies 154 retained category IDs with unique common/scientific strings and a matching source taxonomy leaf. All remain candidates; 28 meet metadata-only support requirements in both detector groups. The prediction unit is a source image-category annotation, not a localized animal. Selection $S$ means at least one publisher-provided MDv4 animal detection has native signed confidence strictly greater than 0.8; the remaining frames form $R$. The cache covers all 201,399 training frames, retaining negative suppression-score signs. Detector scores define selection metadata, not animal-presence truth.

A fixed sequence-ID hash partitions approximately 20\% development and 80\% confirmation with no shared sequences; sites can overlap. Before naming outcomes, each target class must have, in each filter group, 20 total sequences, four development sequences, and ten confirmation sequences. Development contains 885 fixed-hash frames, capped at 16 per class/group, from a 20,139-frame target population. Confirmation independently samples $m_{cg}=\min(128,N_{cg})$ frames without replacement per class/filter stratum: 6,740 frames from 81,969 target frames, with 4,179 sequences and 206 sites. Download failures are retried at the same IDs without redraws.

Rare Species \citep{rarespecies2023} supplies 400 scientific source IDs and 397 native-token-distinct common-name nodes. Before image-model predictions, the image protocol excludes five IDs belonging to two ambiguous common nodes, leaving \RareClasses{} candidates. All 11,983 source images decode; the primary cohort has \RareImages{} rows in \RareGroups{} byte/RGB exact groups, with no cross-class exact conflict. The fixed group-separated split has \RareDevImages{} development images/\RareDevGroups{} groups and \RareConfImages{} confirmation images/\RareConfGroups{} groups. The seven extra rows beyond one per group are retained; this is not seven duplicate-containing groups. All 395 classes have support in both splits. Class-equal summaries describe the complete fixed confirmation cohort, without hypothetical iid sampling intervals. This source has no detector strata and does not replicate the iWildCam selection analysis.

Source labels are not independently verified visible-organism truth. One of the five fixed Rare illustrations is a catalogue ledger page carrying an Orinoco Goose category, rather than a visible-bird photograph; another is labelled coral-specimen media (Appendix~\ref{app:qualitative}). These inspectable examples restrict biological interpretations, without quantifying media prevalence or explaining checkpoint rankings.

\subsection{Models, policies, and chronology}
We use general LAION-2B CLIP ViT-L/14 \citep{laionl14}, BioCLIP2 ViT-L/14 \citep{gu2025}, and BioCLIP 2.5 Huge ViT-H/14 \citep{bioclip25huge2026}. We preserve native FP32 checkpoints, tokenizers, and released preprocessing, including 224-pixel resize/center crop. A checkpoint shares image features across policies. Normalized image/text features yield cosine scores over the same ordered candidates; ties use ascending source ID.

The original four templates are $t_0$: \texttt{a photo of a \{name\}.}; $t_1$: \texttt{a photograph of a \{name\}.}; $t_2$: \texttt{a photo showing a \{name\}.}; and $t_3$: \texttt{this is a photo of a \{name\}.} Each takes common C, scientific S, or dual D: \texttt{\{common\} (\{scientific\})}, giving 12 variants. Original iWildCam $t_0$ C/S/D predictions replay exactly by ID. A separate extension uses the official demo's complete ImageNet80 inference-template bank in archived source order \citep{bioclip2demotemplates2026}, with C/S payloads and four fixed C/S ensembles over 80 or the 79 templates excluding the exact $t_0$ overlap (index 39). These 164 policies retain the same image features and candidates. The bank includes artistic, material, toy and context changes; it is not a certified equivalent-paraphrase set or evidence about training prompts.

Four ensembles use four unit text vectors per class: C-ens uses C$_{0:3}$, S-ens uses S$_{0:3}$, Mix-ens uses C$_0$, C$_1$, S$_0$, S$_1$, and D-ens uses D$_{0:3}$. We use the standard CLIP prompt-ensembling recipe \citep[Sections~2.5 and 3.3]{radford2021}: float64 mean, unit normalization, then FP32 cast. Development class-macro accuracy selects one ensemble globally per checkpoint/source; exact ties follow Mix, C, S, D. A single all-model lock precedes confirmation callbacks, and all policies are reported. Four prototypes equalize count among ensembles, not tokens, architectures, FLOPs, or wall time. The source audit supplies no second verified common alias across the vocabulary; same-name wording does not fill that gap.

Table~\ref{tab:chronology} centralizes outcome access. ``Revision'' denotes outcome-informed analyses, not a claim of fresh confirmation on reused iWildCam frames. Source pairing establishes archived correspondence, not certified semantic equivalence or unchanged taxon scope.

\section{Reviewed wording and prospective confirmation protocol}
\label{app:v10-protocol}
\paragraph{Scope of the human check.}
An AI assistant proposed the first 24-template candidate list as 12 sentence forms crossed with two nouns. One human author was asked to read both noun variants of every form and mark retain, edit, non-equivalent, or uncertain. The rubric was the intended statement that a photograph depicts the named category, without adding quantity, size, visibility, setting, quality, material or a different depiction medium. The author explicitly reported having checked both variants of all 12 forms and retained every candidate. No model-based exclusion or alternative candidate pool preceded this retained list. The candidate list, literal review feedback and finalized text hashes are preserved. The reviewer knew earlier findings and the study question; there is no inter-rater agreement estimate, blind expert certification, or per-species grammatical validation.

Table~\ref{tab:v10-wording} gives the retained wording and the template IDs used in all menu results.

\paragraph{Chronology and distinct families.}
The original naming, four-template, official 80 and detector outcomes were available when this revision began. Equal-sized official-bank menus were enumerated as an exploratory existing-data audit. The new sequence-exclusion rule, 27-class residual population, 5,489-frame draw and three original fresh contrasts were frozen before new image predictions. A separate four-endpoint wording design was fixed before candidate 24 scoring; after human acceptance, all 24 text bytes were frozen before encoding. Existing-data reviewed-pool outcomes then motivated an additional hypothesis about equal-sized, equally noun-balanced menus. The entire 10,626-menu family and its balanced-menu success rule were frozen and mirrored on the server while neither fresh image-encoder output directory nor fresh decision-result directory existed. This is prospective confirmation of a hypothesis suggested by prior data, not a claim that the project had no exploration or was externally preregistered. The three confidence families remain separately identified; we do not pool their nominal 95\% levels into a project-wide guarantee.

\paragraph{New image population and missingness.}
The old 7,625 evaluation IDs occupy 4,835 sequences. Removing all their sequences from the original 81,969-frame confirmation inventory leaves 52,151 frames in 8,100 sequences and 27 classes. Class 12 has no remaining support and is absent from this new target; all 154 candidates still compete. Each supported class uses an independent OS-entropy sample without replacement on sorted original IDs, capped at 256; the exact 5,489-ID draw is reused by every checkpoint. The draw has 3,314 distinct sequences. The original acquisition gate failed on two byte-and-RGB duplicates of old images. Before fresh encoding, a versioned amendment applied the fixed rule ``exclude any ZIP CRC32/uncompressed-size pair present among old evaluation IDs'' to every residual inventory member. Six population entries meet this rule, including two from the original draw. The retained intersection therefore contains 5,487 frames, 3,313 sequences and 27 classes in a 52,145-frame domain. Conditional on each retained class count, the intersection of a class SRS with a fixed metadata domain is SRS in that domain. This is a documented pre-inference amendment, not an unchanged original protocol. CRC collisions may conservatively exclude additional entries; the target is this declared domain, not the unknown set of all distinct biological scenes. No fresh predictions or image judgments determine the screen. Table~\ref{tab:v10-strata} gives all amended class sizes. The acquisition checks ZIP member size/CRC, original bytes, dimensions and stored RGB, and compares exact byte/RGB hashes against all 7,625 earlier evaluated images before allowing inference. The amended retained sample has zero exact old-image overlaps. Download failures are retried at the same ID; there is no redraw or complete-case substitution. \FreshAcquisitionFinding{}

\begin{table}[!htbp]\centering
\tablegraphic{\includegraphics{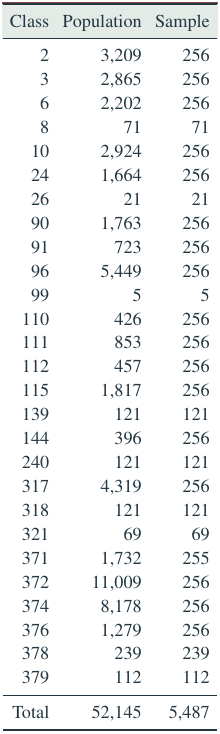}}
\caption{Screened iWildCam frames by class and retained sample size.}
\label{tab:v10-strata}
\end{table}

\paragraph{The seven original fresh endpoints.}
The bank family includes Huge-minus-BioCLIP2 within-common disagreement under the original four templates, under the full official 80, and the paired 80-minus-four change. The expected signs were fixed as positive, negative, negative. The wording family includes within-common disagreement under all 24, photo-only 12, photograph-only 12, and the 12 matched noun substitutions. It specified two-sided reporting without choosing an expected favorable sign. All outcomes use exactly the same new images and candidate IDs. These families have sizes 3 and 4 at separate $\alpha=.05$; the paired change has range width 4, and other checkpoint contrasts width 2.

\paragraph{The all-menu confirmation criterion.}
The additional family contains every lexicographically ordered four-subset of $p_{00},\ldots,p_{23}$. Its success criterion is a positive lower bound and a negative upper bound among the 4,356 subsets with exactly two even/photo and two odd/photograph IDs. Every one of the 10,626 intervals uses the same radius from Equation~\ref{eq:fresh-radius}, with $M=10626$, $R=2$ and the frozen per-class sample sizes. We clip only to the logical range $[-1,1]$. A point sign or an interval containing zero does not certify a reversal. The displayed minimum and maximum balanced-menu estimates are chosen only as deterministic witnesses; ties follow lexicographic order. Because the bounds cover every menu, their simultaneous guarantee includes these displayed witnesses. The complete menu table retains unresolved and opposite cases.

\section{Calculations for prompt and frame budgets}
\label{app:v10-menu-moments}
Equation~\ref{eq:menu-mean} follows because every pair in a uniform $k$-subset has the same inclusion probability. For an exact variance, let $z_e=d_e-\mu$ be each centered pair contrast in the full $n$-template pool, with $\mu=\binom n2^{-1}\sum_e d_e$. Define $S_2=\sum_e z_e^2$, $S_3$ as the sum of $z_ez_f$ over unordered distinct edge pairs sharing a template, and $S_4$ as the sum over disjoint edge pairs. With $\pi_j=\binom kj/\binom nj$ (zero if $k<j$),
\[
 \operatorname{Var}_B D(B)=
 \frac{\pi_2 S_2+2\pi_3 S_3+2\pi_4 S_4}{\binom k2^2}.
\]
This accounts for shared templates; the template-pair values are not treated as independent observations. Direct exhaustive checks for $k=2,4,n$ and small unequal-edge fixtures verify the implementation. The official-bank quantity experiment uses $k=2,4,8,16,32,64,80$. Distributional summaries are exhaustive for 2, 4, 80 and use 10,000 fixed-seed uniform subsets for the remaining sizes; the displayed mean and SD formula is exact for every size. A zero count in a Monte Carlo sign column is not a zero-probability claim. The all-four reviewed-pool distributions are exhaustive, including every noun-composition stratum.

\paragraph{Frame-budget replay.}
For Section~\ref{sec:rq4}, we used NumPy PCG64 with seed 20261122. Within each of 200 repetitions we permuted the original frame rows independently for each class in ascending class-ID order and used nested prefixes. The class means of the 276 reviewed pair contrasts suffice to reconstruct every four-menu estimate. For each cap we applied Equation~\ref{eq:fresh-radius} with $M=10{,}626$, $R=2$ and the actual retained class counts. At the full budget the replay recovers every balanced-menu estimate and the original 43 positive and 1,526 negative decisions. The caps, seed and retrospective analysis are separate from the prospective protocol.

\section{Reproducing the new-sequence decisions}
\label{app:v10-repro}
\FreshReproducibilityFinding{}
The old-image reviewed-pool extension has 1,555,872 retained class decisions. A separate implementation reconstructs every decision from the native features, enumerates pair indicators independently of the producer's frequency-count calculation, and reproduces 28,052 numerical fields with maximum absolute discrepancy $2.23\times10^{-15}$. This is computation verification on reused data, not new-image confirmation. All original V9 numerical results remain archived; the following sections preserve their scope and earlier reproducibility counts. Those historical counts are not totals for the expanded V10 release.

\paragraph{Cross-platform numerical sensitivity.}
A separate macOS arm64 test (Python 3.13.5, NumPy 2.3.3, OpenBLAS 0.3.29) fails the strict exact-prediction check: two of 2,370,384 fresh decisions differ from the Linux production output. Both are Huge near-boundary cases, one scientific-name decision in the official bank and one common-name decision in the reviewed pool. Re-evaluating just their competing stored vectors in FP64 gives score gaps of $1.54\times10^{-7}$ and $1.08\times10^{-7}$. This diagnostic does not replace the frozen FP32 scoring rule. The largest change among the seven reported estimates is 0.001206 pp; all balanced-menu support counts and both displayed witness IDs remain unchanged. We preserve the original outputs, record the portability failure, and scope exact fresh feature rescoring to the supplied Linux numerical environment rather than claiming architecture-independent argmax identity. Conversely, moving the old-image analysis from its original macOS stack to Linux changes one of 1,555,872 decisions; its largest aggregate change is 0.001076 pp. The release therefore separates portable statistical replay from frozen decisions, strict feature rescoring in each original stack, and full raw native encoding. A pass in one route is not reported as a pass in another.

\section{Measurements}
\label{sec:measurements}
For policies $A,B$ and source label $y_i$, define $u_i=\mathbf1[p_i^A\ne p_i^B]$, $v_i=\mathbf1[(p_i^A=y_i)\mathbin{\mathrm{xor}}(p_i^B=y_i)]$, $e_i=\mathbf1[p_i^A\ne y_i,p_i^B\ne y_i]$, and $w_i=e_i u_i$. Both correct, A-only, B-only, same wrong, and different wrong partition the outcomes, giving $u=v+w$. These are accounting identities.

For iWildCam, with $C=28$, $q_c=N_{cR}/(N_{cS}+N_{cR})$, and sampled stratum mean $\bar z_{cg}$,
\begin{align}
\widehat Z_S&=C^{-1}\sum_c\bar z_{cS},\\
\widehat Z_F&=C^{-1}\sum_c[(1-q_c)\bar z_{cS}+q_c\bar z_{cR}].
\end{align}
The target bias is $B_z=Z_S-Z_F$. Development and confirmation restore their own original stratum frequencies, rather than using raw stratified sample fractions. Rare uses equal class weights on its fixed cohort. In each view, $E=\widehat e$, $W=\widehat w$, and $r=W/E$ for positive $E$; this is a ratio under the same weighted measure, not a mean of class ratios.

The original signed outcome is $\delta_i=\mathbf1[p_i^C=y_i]-\mathbf1[p_i^S=y_i]$. Simultaneous design bounds condition on finite-frame outcomes under independent stratum samples, with simple random sampling without replacement within each stratum \citep{hoeffding1963}; they do not assume independent burst outcomes. Original three/supplemental six/nine consistency families remain separate. Five pp is our original audit reporting margin, set by the archived design's \nolinkurl{decision_rule.material_bias_threshold}; it has no demonstrated ecological-utility mapping. A separate revision family covers 522 fixed $E/W$ means and propagates to $r$. Earlier-audit checkpoint comparisons use a post-hoc 279-contrast family: three model pairs, three S/R/F views, and 31 old-template, new-bank and bank-change summaries. It was defined after the old, bank and literal-word macro results were known. None of these families corrects endpoint selection or supplies project-wide coverage. Rare is a fixed-cohort census. Appendix~\ref{app:inference} specifies the bounds and original protocol; the earlier 171 family remains historical.

Let $U_{CS}=\frac14\sum_t\widehat u(C_t,S_t)$ and $U_C,U_S$ average the six unordered within-name template pairs. The diagnostic excess $X=U_{CS}-(U_C+U_S)/2$ compares these fixed perturbations. Their linguistic strengths are not randomized or matched, and shared policies/images are not independent repeats. Error-base and margin diagnostics use the original $t_0$ pair, not these averages.

The external-bank summaries use the same definitions over 80 matched cross-name pairs and all 3,160 within-name pairs, or 79/3,081 after excluding the old default. All within-name pairs are evaluated; they are not independently sampled repetitions. Four-, 79- and 80-prototype ensembles differ in text-computation cost and perturbation coverage.

For a channel, write $\Delta_B=U_B^{\rm Huge}-U_B^{\rm Bio2}$ for menu $B$ and $I_B=\Delta_B-\Delta_4$ for its change from the original four-template menu. This paired menu-by-checkpoint contrast is a functional comparison, not a causal training interaction.

For a concrete choice example, let $A(C_0)$ be common-default source-label accuracy and define $R_C=1-\max_{t<t'}\widehat u(C_t,C_{t'})$ and $R_{CS}=1-\max_t\widehat u(C_t,S_t)$. These are worst-pair agreements over the six/four specified pairs, with the maximum taken after population aggregation. They are neither per-image all-template unanimity nor robustness to unknown templates. A constant predictor can have perfect agreement, so these measurements are not standalone recognition quality. Appendix~\ref{app:notation} collects notation.

\section{Menu-dependent naming and wording comparisons}
\begin{figure*}[!tp]\centering
\includegraphics[width=\textwidth]{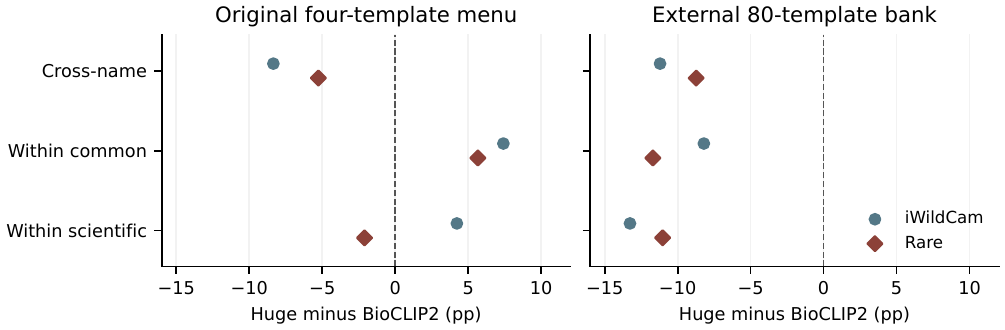}
\caption{Huge-minus-BioCLIP2 differences (pp) for both fixed menus and sources. Panels show point estimates with different perturbation coverage. The external bank includes media/context changes, not 80 equivalent paraphrases. Separate iWildCam 279-family bounds are in Table~\ref{tab:bank-bounds}; there is no causal interpretation.}
\label{fig:checkpoint-profiles}
\end{figure*}

\begin{table}[!htbp]\centering
\tablegraphic{\includegraphics{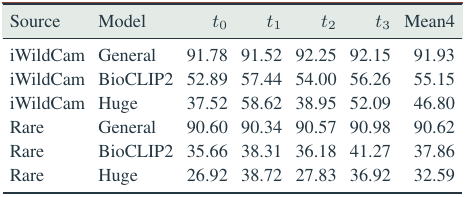}}
\caption{Common/scientific disagreement (\%) for each original template.}
\label{tab:prompt}
\end{table}
\paragraph{An average profile is not a per-template ordering.}
On iWildCam, BioCLIP2/Huge mean cross-name disagreement is \PromptBioCross/\PromptHugeCross\%, while mean within-common rates are \PromptBioWithinC/\PromptHugeWithinC\%. The corresponding Rare rates are \RareBioCross/\RareHugeCross\% and \RareBioWithinC/\RareHugeWithinC\%. Thus the four-template mean and six-pair mean give different checkpoint orderings (Figure~\ref{fig:checkpoint-profiles}). However, $t_1$ (\texttt{a photograph of a \{name\}.}) reverses the cross-name point ordering on both sources: BioCLIP2/Huge are 57.44/58.62\% on iWildCam and 38.31/38.72\% on Rare. This counterexample limits the claim to the specified average. It does not identify training-prompt exposure as its cause.

Within the 279-contrast iWildCam family, Huge-minus-BioCLIP2 old mean cross/within-common differences are $-8.35$ pp $[-14.37,-2.32]$ and $+7.41$ pp $[1.39,13.44]$. The $t_1$ difference is $+1.18$ pp $[-4.84,7.21]$: its point reversal is not an established population-level positive difference. These are separate post-hoc-family bounds, not preregistered tests or joint coverage of all historical analyses.

Within-scientific ordering is source-specific. Huge is above BioCLIP2 on iWildCam (\PromptHugeWithinS\% versus \PromptBioWithinS\%) but below it on Rare (\RareHugeWithinS\% versus \RareBioWithinS\%). We cannot attribute that contrast to source composition, scaling, adaptation, or training: checkpoints and source populations differ on several factors. General CLIP's much higher cross-name disagreement provides an additional reported profile, rather than a checkpoint omitted for its low scientific-name accuracy.

\paragraph{Post-hoc URL metadata sensitivity.}
After the full Rare outcomes and five examples were seen, a fixed URL rule and metadata-only support gate defined \URLClasses{} source classes with at least five images in each URL stratum. The original 395-class analysis remains unchanged. On the same supported classes, Huge-minus-BioCLIP2 cross-name differences are \URLSgroupCrossDelta/\URLOtherCrossDelta{} pp and within-common differences \URLSgroupWithinCDelta/\URLOtherWithinCDelta{} pp in the two strata (Appendix~\ref{app:url}). These directions occur in both groups, but URL strings do not certify photographs or visibility, eliminate source confounding, or provide fresh confirmation.

For the original four-template controls, positive excesses are not purified semantic effects. On Rare, X is positive in \RareBioPositiveExcessClasses/395 BioCLIP2 and \RareHugePositiveExcessClasses/395 Huge class point estimates. The checkpoint contrast is heterogeneous: Table~\ref{tab:class-sensitivity} reports all six source/channel median, IQR and leave-one-class-out summaries. The cross-mean and within-common mean signs persist under deleting any one supported class, without implying random-taxon inference. Full class rows, including opposite directions, remain in the numerical artifact (Appendix~\ref{app:class}).

\paragraph{The broader bank reverses the within-common ordering.}
Table~\ref{tab:external80} changes the perturbation menu while retaining image vectors and candidate labels. BioCLIP2/Huge within-common disagreement is now 59.12/50.91\% on iWildCam and 49.36/37.66\% on Rare; Huge is lower, reversing the original four-template ordering on both sources. Mean cross-name disagreement remains lower for Huge: 72.61/61.40\% and 60.12/51.40\%. Both directions persist after excluding the old default (Appendix~\ref{app:external79}). Thus the original cross/within-common dissociation is a local property of the four-template menu, not a bank-invariant checkpoint characterization. We do not infer why the broader, non-equivalent perturbations change that ordering.
\begin{table}[!htbp]\centering
\tablegraphic{\includegraphics{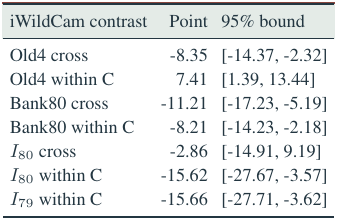}}
\caption{iWildCam bank and bank-change contrasts (pp), exploratory 279-contrast family.}
\label{tab:bank-bounds}
\end{table}
The within-common iWildCam contrast changes by $I_{80}=-15.62$ pp $[-27.67,-3.57]$, and by $I_{79}=-15.66$ pp $[-27.71,-3.62]$. Both change bounds lie below zero in this nominal post-hoc family; they identify neither a training mechanism nor a random-template population. The cross-channel change remains uncertain (Table~\ref{tab:bank-bounds}).
\begin{table*}[!tp]\centering
\tablegraphic{\includegraphics{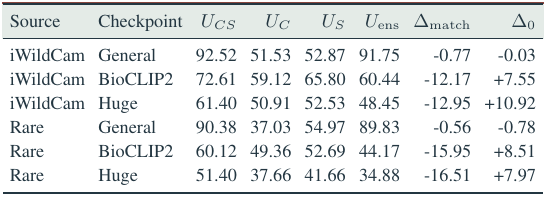}}
\caption{Disagreement (\%) and ensemble changes (pp) for the 80-template bank.}
\label{tab:external80}
\end{table*}
The 80-template C/S ensemble pair also retains reference dependence. BioCLIP2/Huge disagreement is 60.44/48.45\% on iWildCam and 44.17/34.88\% on Rare: lower than the bank's matched-template mean, but $+7.55/+10.92$ and $+8.51/+7.97$ pp above the original $t_0$ pair. These results do not make the bank a universally better policy or imply that all ensemble accuracy changes are unfavorable (Appendix~\ref{app:external79}).

\paragraph{Post-hoc literal template-word sensitivity.}
After the bank means were known, a fixed literal \texttt{photo} word rule partitioned 43/37 templates before cell outcomes were read. The within-common Huge-minus-BioCLIP2 contrast remains negative in photo/photo pairs: $-4.03$ pp on iWildCam and $-0.80$ pp on Rare; other/other and mixed pairs are also negative. This partition therefore does not recover the old four-template direction. It is neither expert semantic coding nor a causal account of perturbation strength (Appendix~\ref{app:photo-word}).

Accuracy tells a different part of the story (Table~\ref{tab:default}). BioCLIP2's iWildCam defaults are \BioCommon\% and \BioScientific\%, yet \BioDisagree\% of decisions differ. Rare instead has a scientific-default accuracy advantage: \RareBioScientificAdvantage{} pp for BioCLIP2 and \RareHugeScientificAdvantage{} pp for Huge. Thus ``nearly equal accuracy'' cannot characterize naming policies universally. Source-reference text retrieval also remains distinct: BioCLIP2 retrieves 129/154 paired scientific names and Huge 141/154; 25 of the 28 image-target BioCLIP2 pairs are mutually Top1-retrieved. Lexical retrieval does not certify interchangeable image responses or biological equivalence (Appendix~\ref{app:lexical}).

\section{Fixed evaluation choices and detector selection}
\subsection{A concrete, descriptive model-choice tradeoff}
\begin{table}[!htbp]\centering
\tablegraphic{\includegraphics{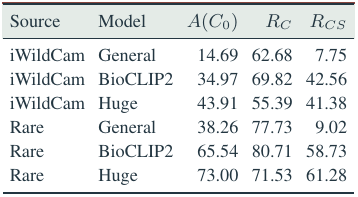}}
\caption{Default common-name accuracy and worst-pair agreement (\%).}
\label{tab:choice}
\end{table}
Table~\ref{tab:choice} makes a local evaluation consequence explicit. Among the three checkpoints, maximizing default-common source-label accuracy favors Huge on both sources: 43.91\% versus BioCLIP2's 34.97\% on iWildCam, and 73.00\% versus 65.54\% on Rare. Maximizing worst common-wording pair agreement over the original four templates instead favors BioCLIP2: 69.82\% versus 55.39\%, and 80.71\% versus 71.53\%. The broader-bank reversal warns against extrapolating that particular tradeoff to another menu; mean and worst-pair summaries are themselves distinct. No outcome-chosen accuracy floor or utility weight is imposed.

Even ``worst agreement'' needs a perturbation set: worst matched naming agreement $R_{CS}$ favors BioCLIP2 on iWildCam but Huge on Rare. None of these summaries establishes ecological benefit, expert correctness or an unknown-template guarantee. A choice requires an externally justified objective; our example shows how explicit objectives change this closed-menu comparison.

\subsection{Development-selected standard ensembles}
\begin{table*}[!tp]\centering
\tablegraphic{\includegraphics{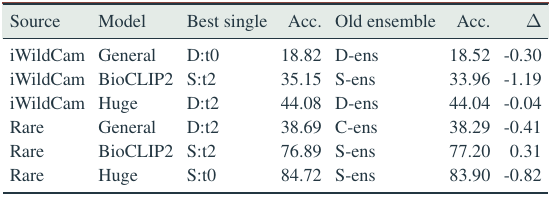}}
\caption{Confirmation accuracy (\%) of development-selected policies. $\Delta$ is ensemble minus single (pp).}
\label{tab:selection}
\end{table*}
The original selection was restricted to four ensembles, so a revision adds best-single and joint-16 development menus while retaining the old choices. All six joint-16 choices are singles (Table~\ref{tab:selection}). Relative to the best-single development choice, the old ensemble's confirmation accuracy changes by $-0.30/-1.19/-0.04$ pp on iWildCam and $-0.41/+0.31/-0.82$ pp on Rare for General/BioCLIP2/Huge. The Rare BioCLIP2 exception precludes saying that ensembles uniformly lose. Development selection need not maximize confirmation accuracy.

The new lock is written after all development calculations but before its confirmation callbacks; the historical confirmation results were already known, so it does not restore fresh blinding. Full menu scores, same-payload comparisons, original $C_0/S_0$ references and all 16 policies remain in the artifact. For example, Rare's original BioCLIP2/Huge S-ens choices are below $S_0$ but above $C_0$; changing payload cannot isolate an ensembling gain (Appendix~\ref{app:policies}).

The original four-template C-ens/S-ens pair is separate from development selection. On Rare its BioCLIP2/Huge disagreement is \RareEnsembleBioU/\RareEnsembleHugeU\%; changes from the matched mean are \RareEnsembleBioDeltaMatched/\RareEnsembleHugeDeltaMatched{} pp, but from $t_0$ they are \RareEnsembleBioDeltaDefault/\RareEnsembleHugeDeltaDefault{} pp. Both reference directions also occur on iWildCam (Table~\ref{tab:ensemble-pair}). Standard averaging therefore retains cross-name decision differences in these evaluated configurations. This does not establish ecological utility or a general failure of prompt ensembling.

\subsection{Original selection result and error base}
\begin{table}[!htbp]\centering
\tablegraphic{\includegraphics{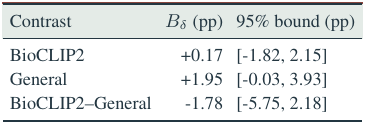}}
\caption{Selection bias in the original three signed iWildCam contrasts (pp).}
\label{tab:primary}
\end{table}
Within-checkpoint signed-bias intervals for BioCLIP2 and General lie inside $[-5,5]$ pp: $[-1.82,2.15]$ and $[-0.03,3.93]$. They exclude five-point signed biases under the original rule but permit nonzero effects, and do not establish stability of individual labels, $W$ or $r$. The BioCLIP2-minus-General interval $[-5.75,2.18]$ excludes a positive five-point contrast but not a negative one; absolute equivalence remains inconclusive. Huge's $[-1.57,2.67]$ interval belongs to a separate supplemental family. None identifies a causal detector effect.

The original nine-family $W$ bias is lower after selection for BioCLIP2 ($-8.79$ pp, $[-9.89,-7.69]$) and Huge ($-6.96$, $[-8.06,-5.86]$). Since $W=Er$, its symmetric arithmetic decomposition is
\begin{align}
W_S-W_F={}&\tfrac12(E_S-E_F)(r_S+r_F)\nonumber\\
 &+\tfrac12(r_S-r_F)(E_S+E_F).
\end{align}
For BioCLIP2 its terms are \ErrorBioBase{} and \ErrorBioConditional{} pp; for Huge, \ErrorHugeBase{} and \ErrorHugeConditional{}. The conditional differences have bounds \ErrorBioRCI{} and \ErrorHugeRCI{} pp; General's is $[-10.65,8.87]$. They include zero and allow large positive and negative changes. Thus neither conditional equivalence nor absence of a conditional effect is established. The point decomposition does not identify a mediated fraction or prove that error-base changes fully explain $W$. Appendix~\ref{app:inference} retains all separate families.

These results motivate three \emph{reporting checks}: show $E,W,r$ with support; distinguish naming from same-name wording controls under explicit references; retain the full candidate vocabulary. They are recommendations, not validated improvements in decision quality. For example, the original $t_0$ audit finds a global opossum-column replacement changes \DistractorBioFlip\% of BioCLIP2 responses despite no own-class target images. The complete margin/replacement diagnostics remain in Appendix~\ref{app:replacement}; their identities are not a mechanism contribution.

\section{Notation and artifact scope}
\label{app:notation}
\begin{table*}[!tp]\centering
\tablegraphic{\includegraphics{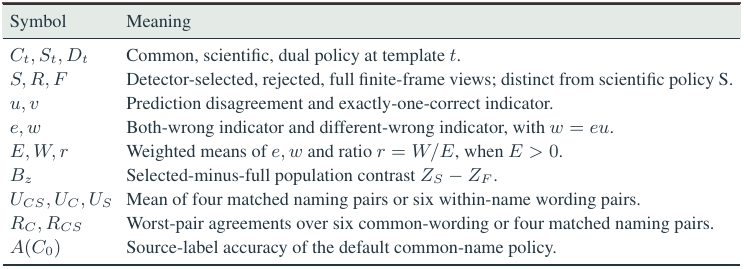}}
\caption{Notation for population views, prediction policies and disagreement.}
\end{table*}
\begin{table*}[!tp]\centering
\tablegraphic{\includegraphics{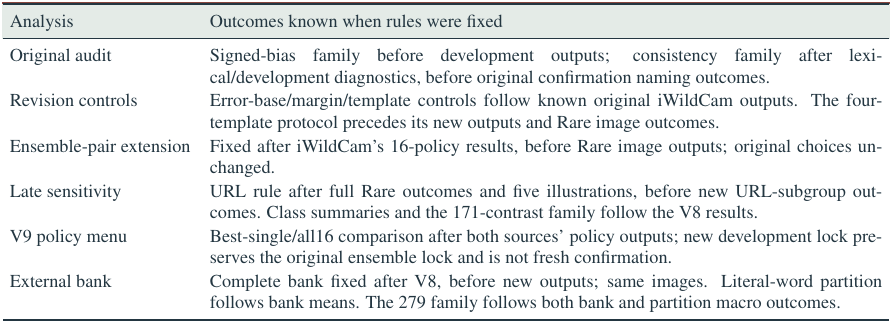}}
\caption{Results available when each analysis was specified.}
\label{tab:chronology}
\end{table*}

All class-level values are preserved as downloadable numerical CSV artifacts; the manuscript replaces the 395-row rendered block with compact distribution summaries rather than removing classes. The source bundle contains \texttt{sources/all\_class\_controls.csv}, \texttt{class\_contrasts.csv}, \texttt{leave\_one\_class\_out.csv}, and the complete paired-bound and policy-selection CSVs. Class deletion changes only the summary's target classes; all candidate labels and image predictions remain unchanged. No leave-out range is a confidence interval for a randomly sampled taxon population.

\section{All fixed iWildCam policy results}
\label{app:policies}
C, S and D denote common, scientific and dual payloads; subscripts denote the four templates. S/R/F mean selected, rejected, and full-population class-macro views. G/B/H abbreviate General/BioCLIP2/Huge. The new confirmation-named outputs reuse the historical iWildCam frames and are post-review results. The development selection lock precedes their new statistics. The table includes every policy, rather than only development-selected or confirmation-best policies.

\begin{table*}[!tp]\centering
\begin{minipage}[t]{.62\textwidth}\vspace{0pt}

\centering
\tablegraphic{\includegraphics{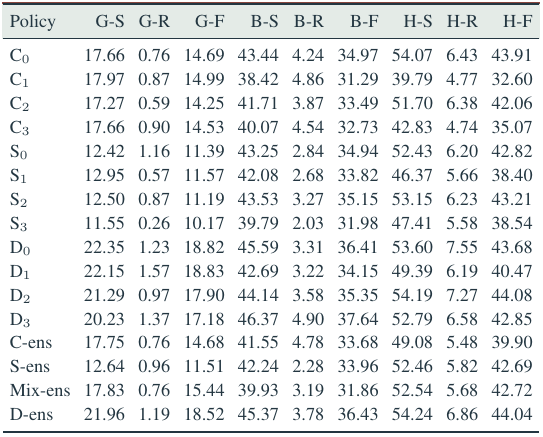}}
\caption{iWildCam confirmation accuracy (\%) by policy, checkpoint and population view.}
\end{minipage}\hfill
\begin{minipage}[t]{.35\textwidth}\vspace{0pt}

\centering
\tablegraphic{\includegraphics{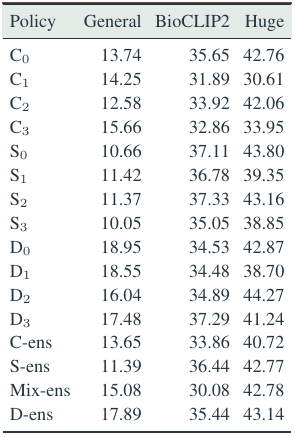}}
\caption{iWildCam development accuracy (\%) over the full population.}
\end{minipage}
\end{table*}
\begin{table*}[!tp]\centering
\tablegraphic{\includegraphics{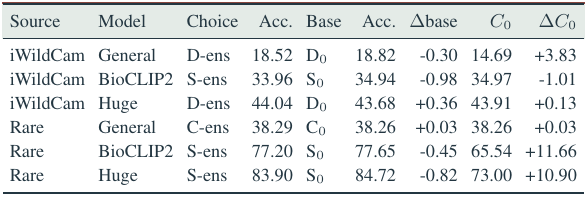}}
\caption{Four-ensemble selection and comparison with the two defaults (\% and pp).}
\end{table*}

\begin{table}[!htbp]\centering
\tablegraphic{\includegraphics{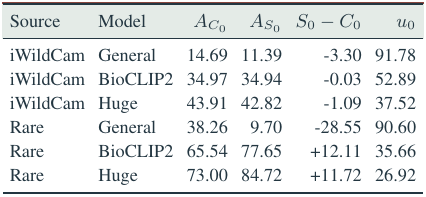}}
\caption{Default source-label accuracy and common/scientific disagreement (\%).}
\label{tab:default}
\end{table}

For each pair below, $u$ is disagreement, $v$ correctness switching, $E$ the both-wrong base, $W$ different wrong predictions, and $r=W/E$. Pair tables are full-population macro percentages. All selected/rejected and per-target-class values are retained in the locked numerical artifact; no favorable pair is omitted.

\begin{table*}[!tp]\centering
\begin{minipage}[t]{.48\textwidth}\vspace{0pt}
\centering
\tablegraphic{\includegraphics{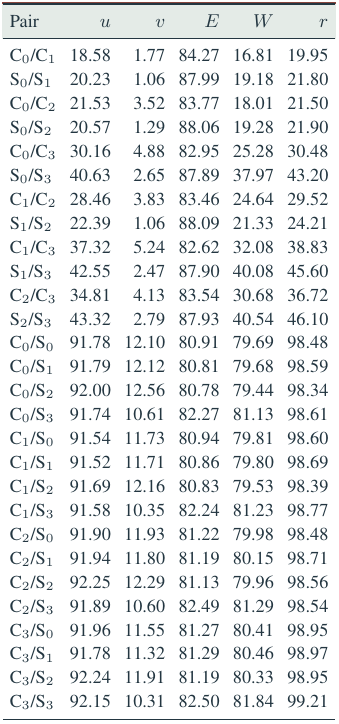}}
\caption{iWildCam pairwise outcomes (\%), general CLIP.}
\end{minipage}\hfill
\begin{minipage}[t]{.48\textwidth}\vspace{0pt}
\centering
\tablegraphic{\includegraphics{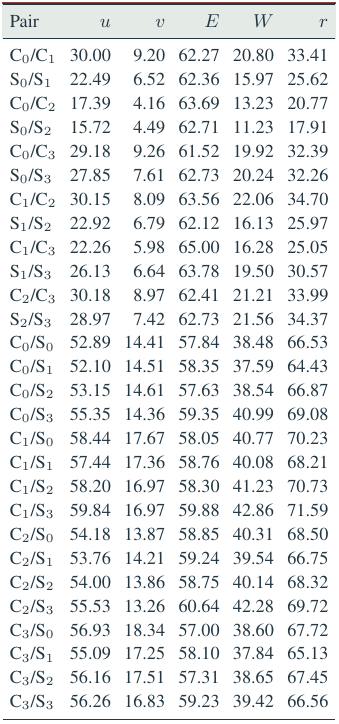}}
\caption{iWildCam pairwise outcomes (\%), BioCLIP2.}
\end{minipage}
\end{table*}

\subsection{Additional fixed ensemble pair}
The one C-ens/S-ens contrast was specified after the parent iWildCam policy outcomes were known. It is not folded into the original families, adds no predictor or policy choice, and is reported for all supported classes and views. The independent numerical replay compared 9,918 scalar fields with zero discrepancies. This is computational verification, not independent image confirmation or human expert annotation.

\section{Original outcomes and inference families}
\label{app:inference}
The original $t_0$ five patterns and all nine consistency contrasts remain below. Their original 95\% intervals were fixed before the original confirmation outputs. They are not replaced by the new post-review error-base bounds.

\begin{table*}[!tp]\centering
\captionsetup{justification=raggedright,singlelinecheck=false,skip=5pt}
\begin{minipage}[t]{.375\textwidth}\vspace{0pt}\raggedright
\tablegraphic{\includegraphics{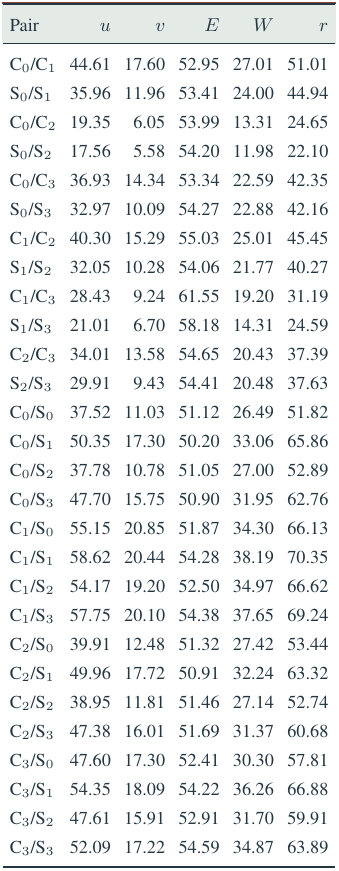}}
\caption{Huge pairs on iWildCam (\%).}
\end{minipage}\hfill
\begin{minipage}[t]{.595\textwidth}\vspace{0pt}\raggedright
\tablegraphic{\includegraphics{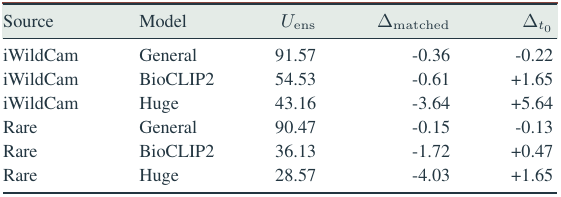}}
\caption{Ensemble disagreement (\%) and changes from the matched and default prompts (pp).}
\label{tab:ensemble-pair}
\par\vspace{12pt}
\tablegraphic{\includegraphics{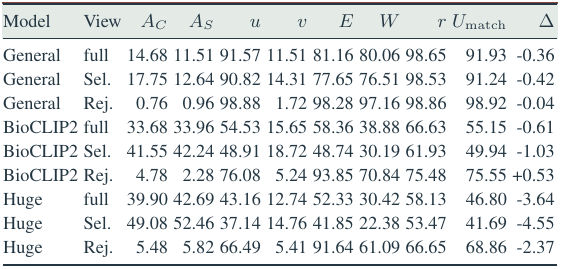}}
\caption{iWildCam ensemble pair by view. Rates in \%, $\Delta=u-U_{\rm match}$ in pp.}
\par\vspace{12pt}
\tablegraphic{\includegraphics{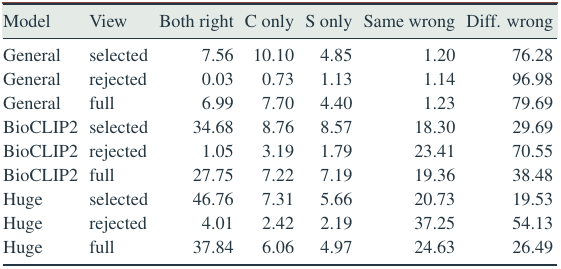}}
\caption{Five paired outcomes on iWildCam (\%).}
\end{minipage}
\end{table*}

\begin{table}[!htbp]\centering
\tablegraphic{\includegraphics{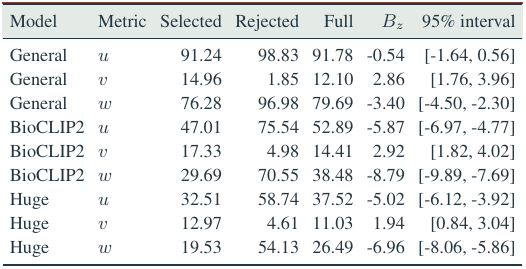}}
\caption{Nine iWildCam consistency contrasts. Rates in \%, selection biases and intervals in pp.}
\end{table}

\begin{table*}[!tp]\centering
\begin{minipage}[t]{.48\textwidth}\vspace{0pt}
\centering
\tablegraphic{\includegraphics{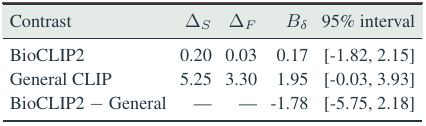}}
\caption{Original signed advantages and selection biases (pp).}
\end{minipage}\hfill
\begin{minipage}[t]{.48\textwidth}\vspace{0pt}
\centering
\tablegraphic{\includegraphics{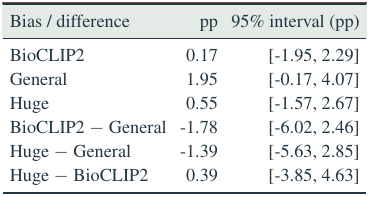}}
\caption{Supplemental signed contrasts and simultaneous intervals (pp).}
\end{minipage}
\end{table*}

For a fixed measurement of range length $L$, a family of $K$ bias contrasts has conservative half-width
\begin{align}
 h_{K,L}&=L\sqrt{\tfrac12 V\log(2K/.05)},\\
 V&=C^{-2}\sum_c q_c^2\sum_{g\in\{S,R\}}
       \frac{\mathbf1[m_{cg}<N_{cg}]}{m_{cg}}.
\end{align}
Census strata contribute zero uncertainty. The original primary family has $K=3$; the expanded supplemental family has $K=6$. Signed correctness has $L=2$, and paired between-model signed-bias differences use $L=4$. The additional binary $u,v,w$ family has $K=9$ and $L=1$. These bounds follow from independent stratum draws and a union bound, without assuming independent finite outcomes inside camera bursts. Their coverage concerns the fixed source-label population.

The post-review family covers 522 binary means: three models, 28 classes plus one class-macro summary, three population views, and $e,w$. For a mean $\widehat Z=\sum_h a_h\bar z_h$ the half-width is $\sqrt{\tfrac12 V_Z\log(2\cdot522/.05)}$, with $V_Z=\sum_{h:m_h<N_h}a_h^2/m_h$. Intersect mean intervals with $[0,1]$. If the $E$ lower bound is positive, propagate to $[W_L/E_U,W_U/E_L]\cap[0,1]$; otherwise use $[0,1]$, while retaining an undefined point ratio when the observed denominator is zero. Differences use endpoints of that same joint event. Nominal coverage is for this fixed post-review family, not conditional on the outcomes that motivated its selection.

The historical five-pp rule is recorded in \nolinkurl{protocol/iwildcam_frame_language_confirmation_design_v2.json}, \nolinkurl{decision_rule.material_bias_threshold}$=0.05$. It is our reporting rule, not a demonstrated ecological cost or a field-standard equivalence threshold. Signed-bias exclusion under this rule does not bound conditional consistency.

\subsection{Direct revision checkpoint families}
\label{app:direct-bounds}
The earlier-audit 279 family comprises three checkpoint pairs, three views and 31 measurements: the old 19 template/pair/mean summaries, six 80/79 bank means and six bank-minus-old interactions. Ordinary per-image checkpoint differences lie in $[-1,1]$, while interactions lie in $[-2,2]$. With fixed population weights $a_h$, the half-width is $L\sqrt{\tfrac12\sum_{h:m_h<N_h}a_h^2/m_h\log(2\cdot279/.05)}$, using range length $L=2$ or $4$ respectively. Census strata contribute zero; there is no iid class-sampling assumption or partial finite-population correction. The family was fixed after all bank and literal-partition macro outcomes, so its nominal coverage is not adjusted for the selection that motivated it. It is separate from every historical family and supplies no project-wide guarantee.

The earlier 171 family below contains only three checkpoint pairs, 19 original-four measurements and three views. It uses the same formula with $K=171,L=2$, and remains unchanged as historical evidence. It does not cover new bank means or interactions and is not the main-text interval source.
\begin{table}[!htbp]\centering
\tablegraphic{\includegraphics{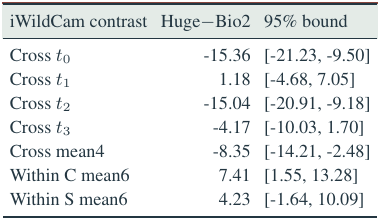}}
\caption{Full-population iWildCam contrasts in the historical 171-contrast family (pp).}
\end{table}

\begin{table}[!htbp]\centering
\tablegraphic{\includegraphics{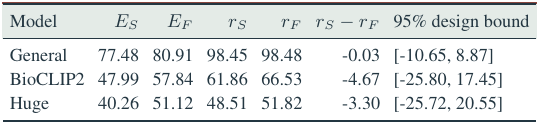}}
\caption{Both-wrong rates (\%) and selected-minus-full conditional disagreement (pp).}
\end{table}
\begin{figure*}[!tp]\centering
\includegraphics[width=.95\textwidth]{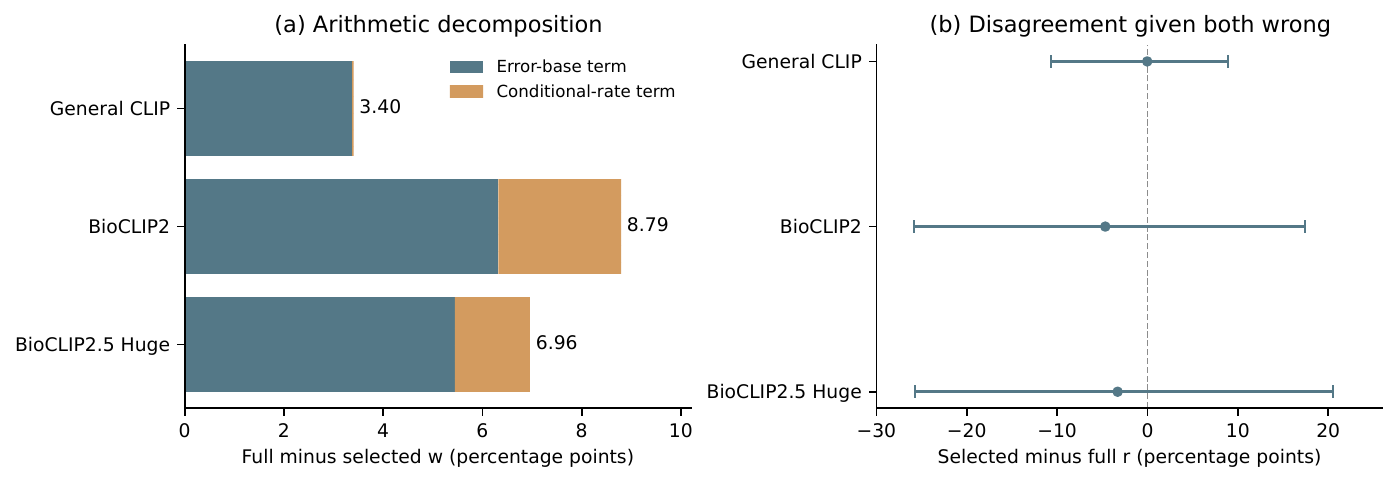}
\caption{Preserved original-$t_0$ post-review diagnostic. Left: $W_F-W_S$ and its symmetric arithmetic terms, opposite in sign to the selected-minus-full differences reported in the body. Right: $r_S-r_F$ with bounds from the separate 522-mean family; every interval includes zero. These are not causal mediation or intervals for the new template averages.}
\label{fig:error}
\end{figure*}
The complete 154 candidate reference list is retained; 126 categories outside the 28-class image target have no own-class image estimate. Zero-denominator classes are reported rather than silently removed. The symmetric product terms are arithmetic attributes of the difference, not estimates of mediation.

To make the uncertainty explicit, BioCLIP2's $r_S-r_F$ bound excludes a $+20$ pp change but permits $-20$ pp; Huge's permits both. Twenty points is merely a numerical reading of the bounds, not a utility threshold or a new test. The selected-minus-rejected contrasts in Table~\ref{tab:conditional-identification} inherit endpoint projection from the same 522 primitive mean intervals; they are not newly tightened or causal effects. They distinguish selection versus rejection from selection versus their population mixture.
\begin{table}[!htbp]\centering
\tablegraphic{\includegraphics{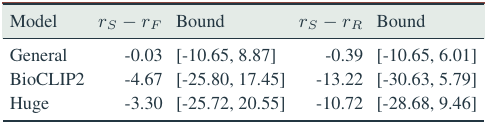}}
\caption{Conditional disagreement contrasts (pp), with bounds from the 522-mean family.}
\label{tab:conditional-identification}
\end{table}

\subsection{Earlier descriptive visual-proxy control}
The original audit also standardized selected/rejected comparisons within development-defined visual-margin bins. Centroids and thresholds were fixed from development features; the true source label chooses the margin, so the diagnostic is an oracle, not deployable confidence. The two image towers are themselves language-pretrained. Shared-bin support changes the estimand, and no adjusted confirmatory intervals were computed. This control neither verifies visibility nor resolves the new conditional-error uncertainty.
\begin{table}[!htbp]\centering
\tablegraphic{\includegraphics{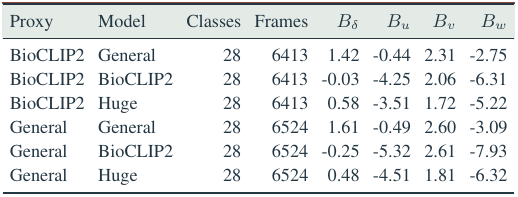}}
\caption{Shared-support iWildCam contrasts standardized by visual proxy (pp).}
\end{table}

\section{Source-reference retrieval and target support}
\label{app:lexical}
Text-only retrieval ranks the other naming-policy prototypes by cosine similarity. MRR uses the first valid positive, and a fixed numeric ID resolves ties. For homonymous common prompts, every corresponding source scientific ID is valid. The iWildCam task uses 154 queries/candidates in each direction; Rare Species has 397 common queries over 400 scientific candidates and 400 reverse queries over 397 common nodes. The latter is the completed lexical task, not the completed 395-class image study.

\begin{table*}[!tp]\centering
\begin{minipage}[t]{.48\textwidth}\vspace{0pt}
\centering
\tablegraphic{\includegraphics{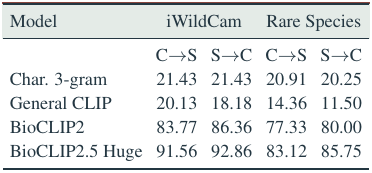}}
\caption{Top-1 text retrieval of the paired source name (\%).}
\end{minipage}\hfill
\begin{minipage}[t]{.48\textwidth}\vspace{0pt}
\centering
\tablegraphic{\includegraphics{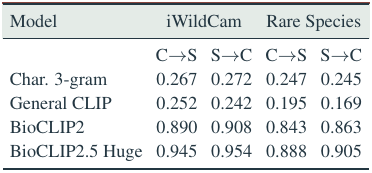}}
\caption{Mean reciprocal rank of the paired source name.}
\end{minipage}
\end{table*}
The character control uses binary NFKC/case-folded character 3-grams with fixed IDF weights. On queries constructed to share no positive-reference 3-gram, its zero accuracy follows from construction and is not an independent semantic control. Exact-string-disjoint slices likewise do not certify biological or pretraining disjointness. Complete per-query ranks and graph links remain in the v7 numerical artifact.

\begin{table}[!htbp]\centering
\tablegraphic{\includegraphics{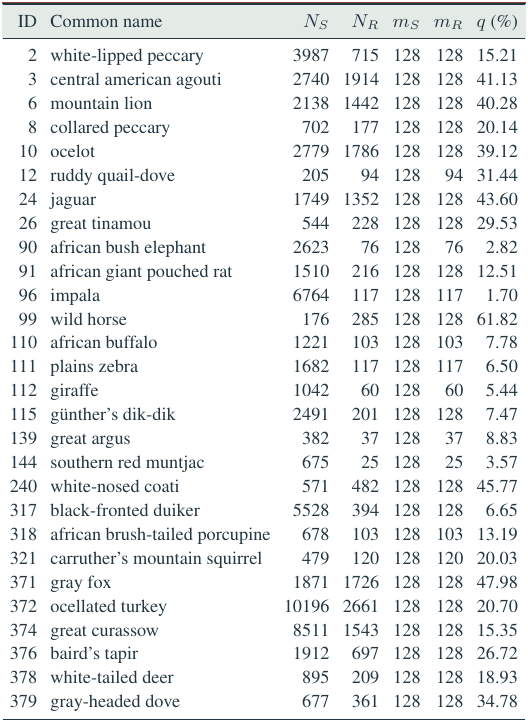}}
\caption{iWildCam confirmation strata. $N$ denotes population size and $m$ sample size.}
\end{table}

\section{Functional candidate and margin audit}
\label{app:replacement}
\subsection{Original \texorpdfstring{$t_0$}{t0} diagnostic results}
\begin{figure*}[!tp]
\centering
\includegraphics[width=.95\textwidth]{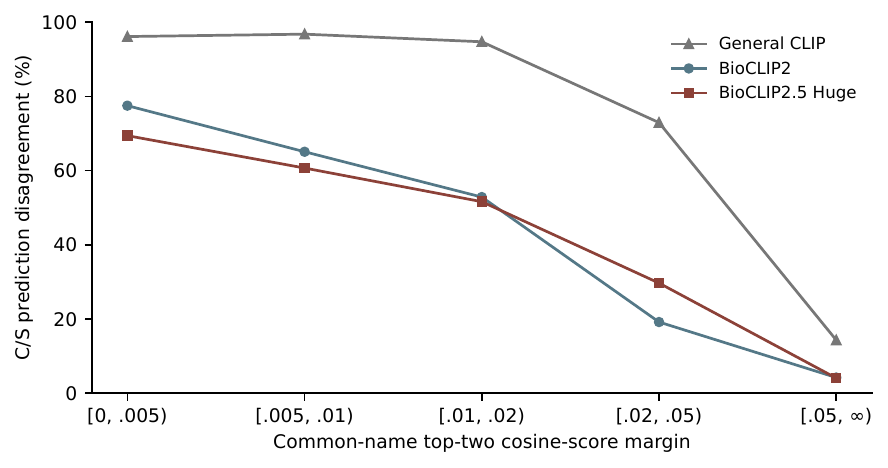}
\caption{Original iWildCam $t_0$ pair: post-review full-population rates in fixed common top1--top2 margin bins. Bins and native cosine origins are fixed; margins are model-specific rather than cross-model calibrated confidence.}
\label{fig:margin}
\end{figure*}
Low common-policy margins concentrate flips. Margins below 0.02 account for \MarginBioLowShare\% of BioCLIP2's weighted flip mass and \MarginHugeLowShare\% of Huge's. But among frames with common margin at least 0.05, the conditional flip rates are still \MarginBioHighRate\% and \MarginHugeHighRate\%. Boundary proximity is informative without reducing every change to an exact tie or a tiny margin.

To connect the endpoints, let $a$ and $b$ be the common and scientific winners, $d_C=s^C_a-s^C_b$, $d_S=s^S_a-s^S_b$, and $\Delta_k=s^S_k-s^C_k$. The ranking identity $d_S=d_C+\Delta_a-\Delta_b$ is verified for every flip and stable frame. It locates score changes without supplying a new theory.

We also evaluate all 154 global one-column C$\to$S replacements and all 154 reverse replacements. Each policy is fixed for every image, preserves all candidates, and does not use $y$ to decide which column to replace. For an illustrative post-review case, renaming candidate 67 from ``gray four-eyed opossum'' to ``philander opossum'' changes \DistractorBioFlip\% of BioCLIP2 decisions and changes accuracy by \DistractorBioDelta{} pp, although no target image has $y=67$. Competing names can therefore matter without changing the true-class prototype. Complete matrices and all policy results, including eight fixed half-candidate masks, are retained; this example is not a selected mitigation. Native-score mixed policies are distinguished from an oracle that selects columns using the unknown label (Appendix~\ref{app:replacement}).

For every one of the 154 candidates, the one-column policy replaces its common prototype with the scientific prototype globally, leaving the other 153 unchanged; the reverse direction starts from the scientific policy. There are 308 policies per model, all evaluated for every image. We retain complete prediction, flip and correctness-change matrices, plus all population/true-class summaries. Evaluations conditional on $y=k$ versus $y\ne k$ use weighted numerators divided by weighted support. A candidate outside the image target has no own-class support, which is explicit rather than assigned zero effect.

Four fixed SHA-ranked 77-candidate masks and their complements add eight global functional controls. They never use labels or outcomes to choose columns. Rankings are descriptive only. They are not ordinary aliases or same-name wording controls, and their effects do not sum to a unique attribution of the full naming contrast.

Hybrid policies depend on their specified native cosine coordinates. For example, a common score row $(.6,.4)$ and scientific row $(.3,.7)$ select different classes; replacing only the first common column gives $(.3,.4)$. Adding .2 to the complete scientific row preserves its winner but changes that hybrid to $(.5,.4)$. This illustrates origin dependence, rather than a failure of the fixed native policy. In particular, a true-class-only hybrid would require $y$ and would be an oracle; our global policies do not choose columns that way.

For the original $t_0$ pair, per-frame files retain common/scientific winners, winner-pair score changes, exact tie flags, true-class margins, strongest wrong competitor IDs, and original population weights. Fixed common top1--top2 bins are $[0,.005)$, $[.005,.01)$, $[.01,.02)$, $[.02,.05)$, and $[.05,\infty)$. Stable frames with $a=b$ are not falsely marked as two-candidate ties. All 154 cross-name cosine/Euclidean prototype displacements are retained. No native token-ID arrays were available in that released feature bundle, so this diagnostic does not infer token equality from strings or vectors. No genus-prefix grouping is treated as verified phylogeny.

\section{All fixed Rare policies and class profiles}
\label{app:class}
The Rare confirmation cohort contains all 395 fixed source classes. The tables below retain every policy and all 28 single-variant naming/wording pairs. The complete 395-class and 28-target class rows are retained in the numerical CSV artifact rather than a long rendered class table. Class-level X values and the following Huge-minus-BioCLIP2 summaries are descriptive, not independent significance tests. G/B/H abbreviate General/BioCLIP2/Huge.
\begin{table*}[!tp]\centering
\tablegraphic{\includegraphics{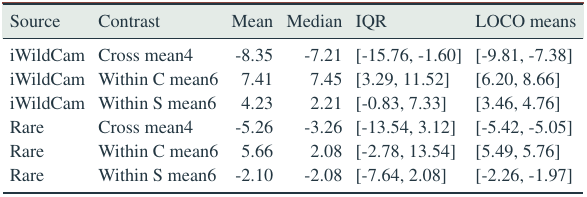}}
\caption{Class contrasts and leave-one-class-out (LOCO) mean ranges (pp).}
\label{tab:class-sensitivity}
\end{table*}
\begin{table}[!htbp]\centering
\tablegraphic{\includegraphics{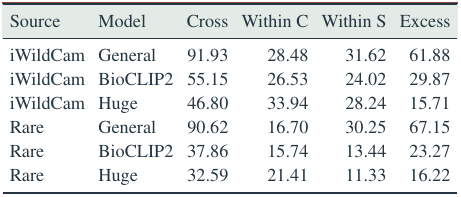}}
\caption{Original four templates. Disagreement rates in \% and excess in pp.}
\end{table}

\begin{table*}[!tp]\centering
\begin{minipage}[t]{.48\textwidth}\vspace{0pt}
\centering
\tablegraphic{\includegraphics{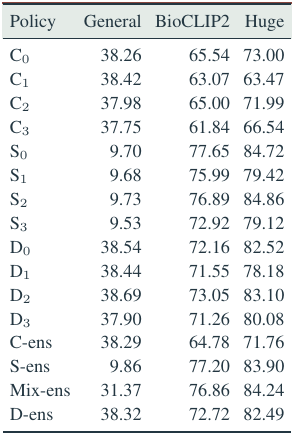}}
\caption{Rare Species confirmation accuracy (\%) for all 16 policies.}
\end{minipage}\hfill
\begin{minipage}[t]{.48\textwidth}\vspace{0pt}
\centering
\tablegraphic{\includegraphics{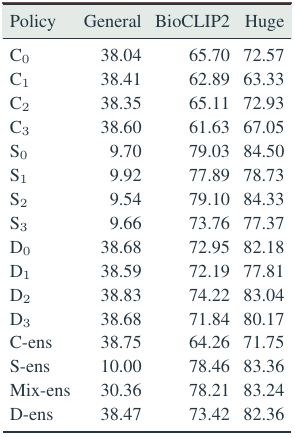}}
\caption{Rare Species development accuracy (\%) for all 16 policies.}
\end{minipage}
\end{table*}

\begin{table*}[!tp]\centering
\begin{minipage}[t]{.48\textwidth}\vspace{0pt}
\centering
\tablegraphic{\includegraphics{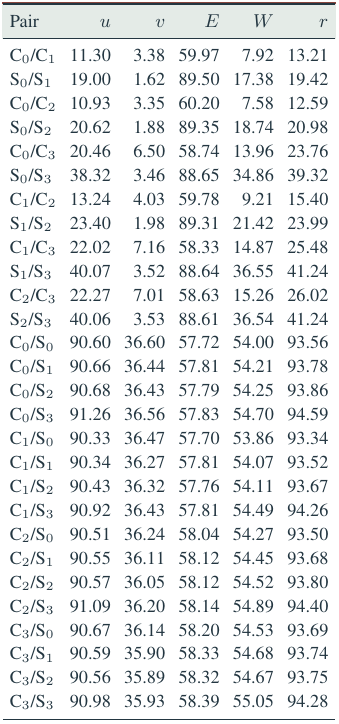}}
\caption{Rare Species pairwise outcomes (\%), general CLIP.}
\end{minipage}\hfill
\begin{minipage}[t]{.48\textwidth}\vspace{0pt}
\centering
\tablegraphic{\includegraphics{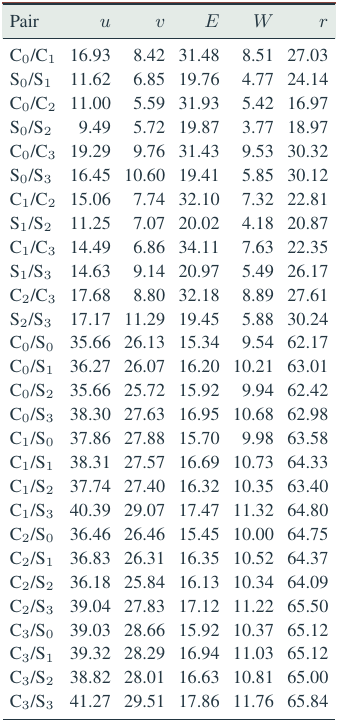}}
\caption{Rare Species pairwise outcomes (\%), BioCLIP2.}
\end{minipage}
\end{table*}


\subsection{External-bank scope and development selection}
\label{app:external79}
The external bank retains all 80 archived templates in source order; index 39 exactly overlaps the old $t_0$. The 79-template sensitivity excludes only that string, without outcome-based deletions. The within-common mean ordering differs from the original four-template menu even in this non-overlap scope. All 160 singles, four ensembles, matched pairs, 3,160 within-name pairs and class-level summaries remain in the artifact.

Best common/scientific singles are selected independently using development class-macro accuracy, with source-order ties; all six source/checkpoint selections are locked before the new confirmation aggregation. Old V8 confirmation outcomes were already known. Table~\ref{tab:external-selection} compares their actual confirmation accuracy with matching-payload 80-vector ensembles. It reports favorable and unfavorable directions rather than claiming uniform ensemble inferiority. The complete new bank scoring uses cached image features and is not fresh image encoding.
\begin{table*}[!tp]\centering
\captionsetup{justification=raggedright,singlelinecheck=false,skip=5pt}
\begin{minipage}[t]{.375\textwidth}\vspace{0pt}\raggedright
\tablegraphic{\includegraphics{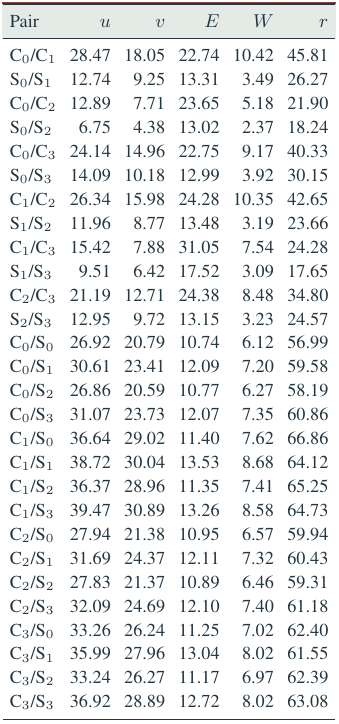}}
\caption{Rare pairs for Huge (\%).}
\end{minipage}\hfill
\begin{minipage}[t]{.595\textwidth}\vspace{0pt}\raggedright
\tablegraphic{\includegraphics{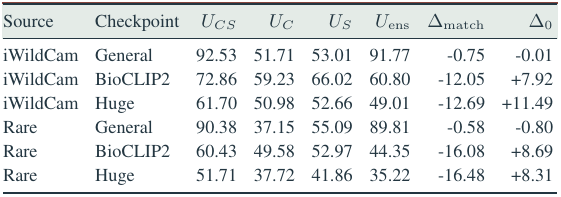}}
\caption{Disagreement in the 79-template bank (\% and pp).}
\par\vspace{12pt}
\tablegraphic{\includegraphics{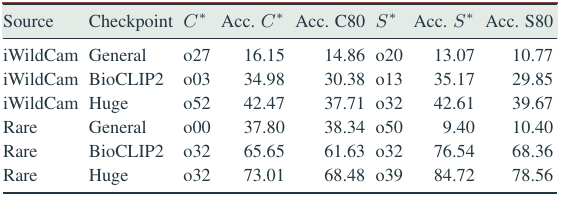}}
\caption{Accuracy of singles and 80-template ensembles (\%).}
\label{tab:external-selection}
\par\vspace{12pt}
\tablegraphic{\includegraphics{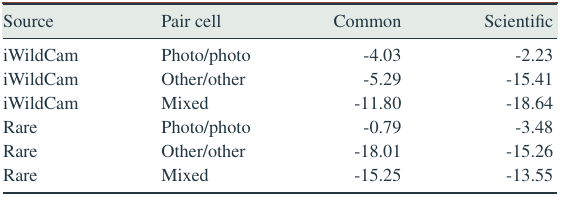}}
\caption{Within-name contrasts by literal \emph{photo} wording (pp).}
\end{minipage}
\end{table*}

\subsection{Literal template-word partition after bank outcomes}
\label{app:photo-word}
The new partition was fixed after all bank macro means were known. A case-sensitive Unicode-word-boundary search for \texttt{photo} in the exact template before name insertion gives 43 photo-word and 37 other-word strings; the latter is not a non-photographic or image-visibility label. The 903 photo/photo, 666 other/other and 1,591 mixed unordered pairs reconstruct all 3,160 within-name pairs with their exact count weights. This arithmetic partition does not change any classifier, candidate, development choice or original cohort.
Complete matched-cross cells, both phases, all views and all three checkpoints remain in the source CSVs. The non-positive within-common direction across these cells does not isolate why the banks differ, equate perturbation strength or verify photographic content.

\section{Post-hoc Rare URL metadata sensitivity}
\label{app:url}
This supplement was fixed after the complete 395-class results and five illustrations were seen, before reading the newly defined subgroup outcomes. A row belongs to S3-URL only when the publisher source URL has scheme \texttt{http} or \texttt{https}, exact hostname \nolinkurl{inaturalist-open-data.s3.amazonaws.com}, and a path beginning \texttt{/photos/}. Every other, missing, malformed or non-string URL belongs to Other-URL. These are metadata strings, not photographic-medium or visibility labels; Other-URL is not a non-photo class.

The original 9,467 confirmation rows contain 3,630 S3-URL and 5,837 Other-URL rows. The pre-outcome coverage condition requires at least five rows per URL group for a class, giving the same $K=316$ classes for all checkpoints; coverage does not guarantee power. Full-on-K retains 7,578 rows, split into 3,476 and 4,102. All rows in each view contribute, first averaging within class and then equally across the same K. All 395 candidates still compete. The other 79 classes retain their labels and remain in the primary analysis.

\begin{table*}[!tp]\centering
\begin{minipage}[t]{\columnwidth}\vspace{0pt}\centering
\tablegraphic{\includegraphics{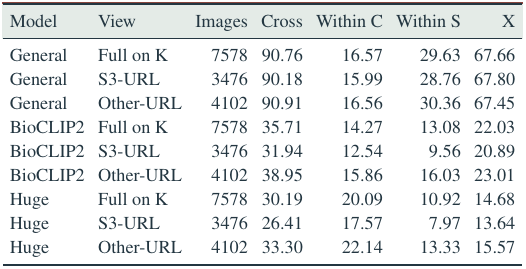}}
\caption{Disagreement (\%) and excess (pp) within 316 Rare classes supported in both URL strata.}
\label{tab:url}
\end{minipage}\hfill
\begin{minipage}[t]{\columnwidth}\vspace{0pt}
Table~\ref{tab:url} summarizes the three URL views. The supplement retains all 252 aggregate pair rows, 79,632 class/pair rows, 2,844 class diagnostics and 96 Huge-minus-BioCLIP2 contrasts. Independent integer-count code reproduced \URLIndependentChecks{} fields to within $10^{-12}$ (maximum error $9.99\times10^{-16}$). This analysis adds no accuracy, $v/w$, policy selection, visibility labels, intervals or $p$-values. Cross-name and within-common signs persist in both strata. Source and medium remain confounded.
\end{minipage}
\end{table*}

\section{Five fixed Rare response illustrations}
\label{app:qualitative}
The unchanged rule partitions the licensed confirmation pool into five BioCLIP2 $t_0$ response cells, then picks the smallest fixed SHA of each image ID. All five selected cells are rendered, with no skips or replacements. Original source bytes and selected IDs remain unchanged after the native-transform compatibility repair. These are response illustrations relative to publisher labels, not a representative error sample, expert identification, media-authenticity certification or an animal-visibility audit. The same-nonmatching example displays a catalogue ledger page; the different-nonmatching example displays labelled coral-specimen media. Stored RGB is shown beside reconstructed native 224-pixel geometry, not a saved historical inference tensor. Complete attribution and clickable source links accompany every panel and figures/rare\_examples/index.html.
\begin{figure}[!htbp]\centering
\includegraphics[width=\linewidth]{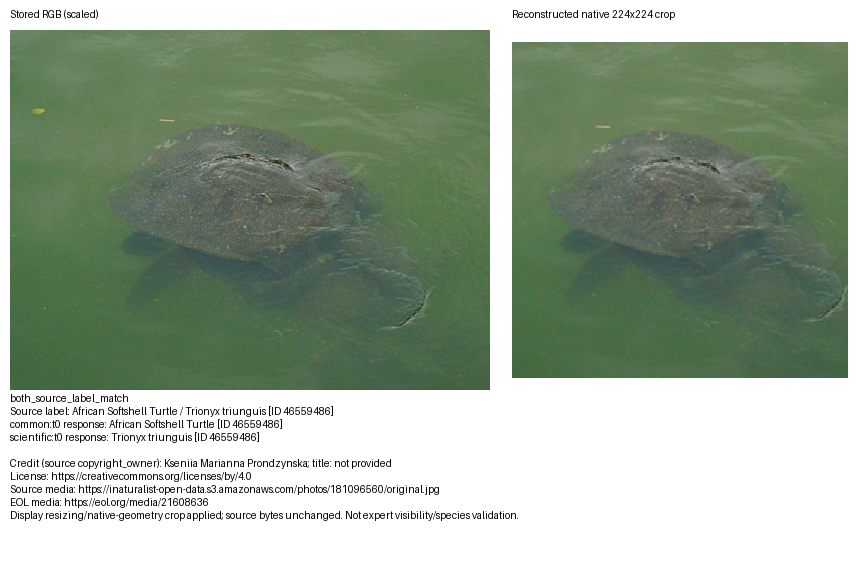}
\caption{both source label match. Source label: African Softshell Turtle / Trionyx triunguis. C-response: African Softshell Turtle; S-response: Trionyx triunguis. Credit (publisher field): Kseniia Marianna Prondzynska. Title: not provided. \href{https://inaturalist-open-data.s3.amazonaws.com/photos/181096560/original.jpg}{Source media}; \href{https://eol.org/media/21608636}{EOL media}; \href{https://eol.org/pages/46559486}{EOL class}; \href{https://creativecommons.org/licenses/by/4.0}{cc-by-4.0}. Display resizing and reconstructed native crop applied; original bytes unchanged. Source-label responses, not expert visibility/species certification.}
\label{fig:rare-both-source-label-match}
\end{figure}
\begin{figure}[!htbp]\centering
\includegraphics[width=\linewidth]{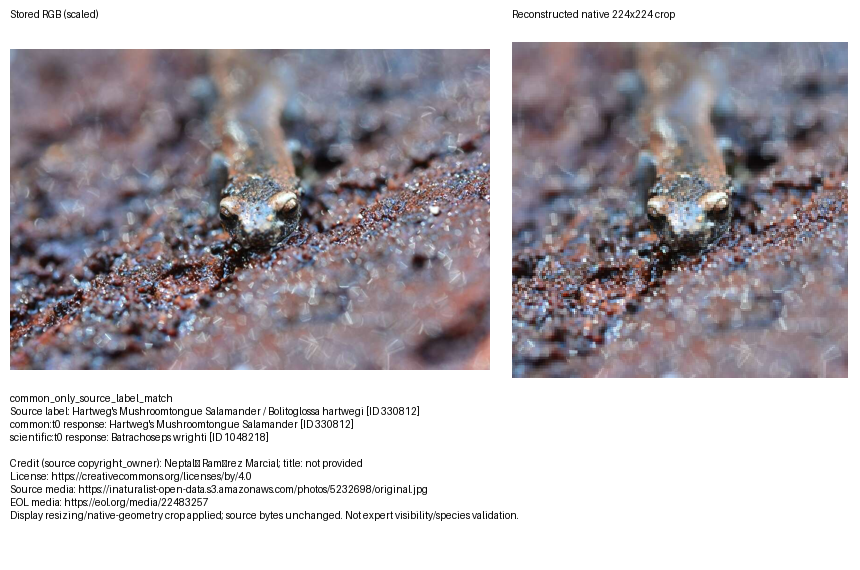}
\caption{common only source label match. Source label: Hartweg's Mushroomtongue Salamander / Bolitoglossa hartwegi. C-response: Hartweg's Mushroomtongue Salamander; S-response: Batrachoseps wrighti. Credit (publisher field): Neptalí Ramírez Marcial. Title: not provided. \href{https://inaturalist-open-data.s3.amazonaws.com/photos/5232698/original.jpg}{Source media}; \href{https://eol.org/media/22483257}{EOL media}; \href{https://eol.org/pages/330812}{EOL class}; \href{https://creativecommons.org/licenses/by/4.0}{cc-by-4.0}. Display resizing and reconstructed native crop applied; original bytes unchanged. Source-label responses, not expert visibility/species certification.}
\label{fig:rare-common-only-source-label-match}
\end{figure}
\begin{figure}[!htbp]\centering
\includegraphics[width=\linewidth]{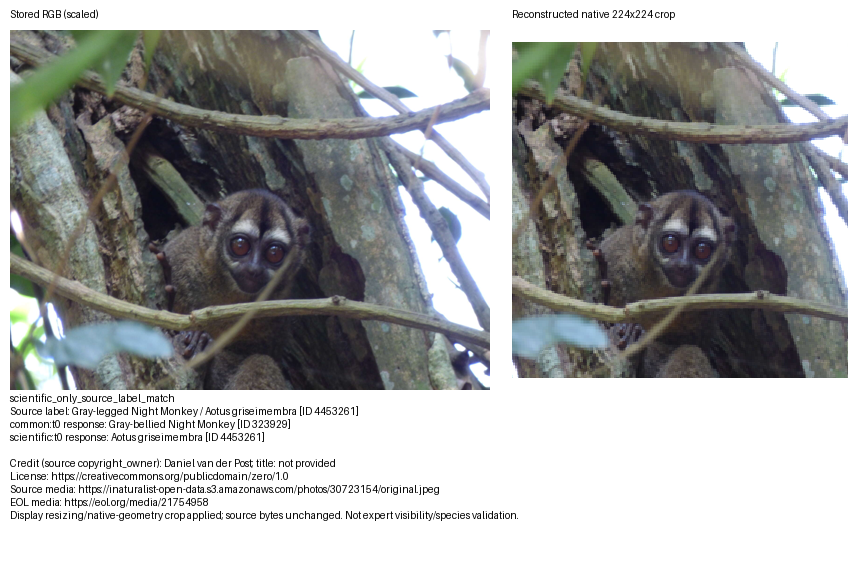}
\caption{scientific only source label match. Source label: Gray-legged Night Monkey / Aotus griseimembra. C-response: Gray-bellied Night Monkey; S-response: Aotus griseimembra. Credit (publisher field): Daniel van der Post. Title: not provided. \href{https://inaturalist-open-data.s3.amazonaws.com/photos/30723154/original.jpeg}{Source media}; \href{https://eol.org/media/21754958}{EOL media}; \href{https://eol.org/pages/4453261}{EOL class}; \href{https://creativecommons.org/publicdomain/zero/1.0}{cc-publicdomain}. Display resizing and reconstructed native crop applied; original bytes unchanged. Source-label responses, not expert visibility/species certification.}
\label{fig:rare-scientific-only-source-label-match}
\end{figure}
\begin{figure}[!htbp]\centering
\includegraphics[width=\linewidth]{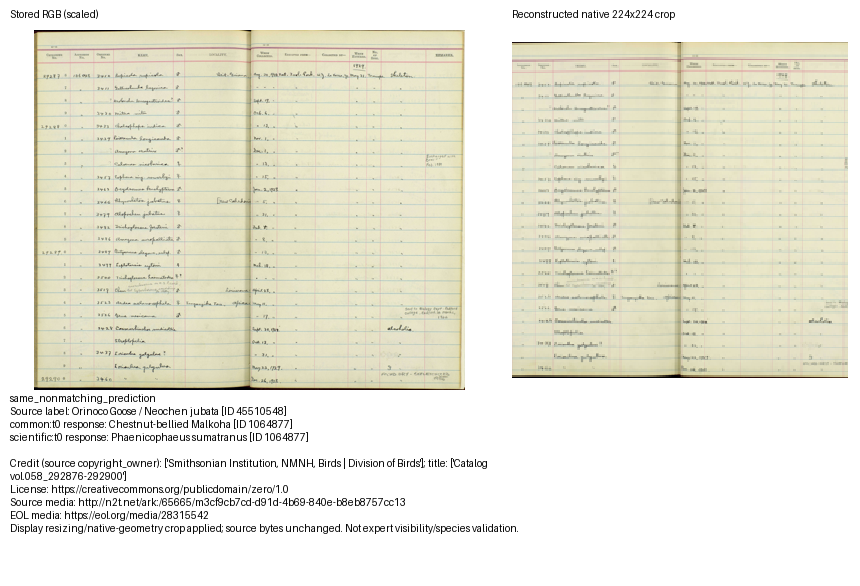}
\caption{same nonmatching prediction. Source label: Orinoco Goose / Neochen jubata. C-response: Chestnut-bellied Malkoha; S-response: Phaenicophaeus sumatranus. Credit (publisher field): ['Smithsonian Institution, NMNH, Birds | Division of Birds']. Title: ['Catalog vol.058\_292876-292900']. \href{http://n2t.net/ark:/65665/m3cf9cb7cd-d91d-4b69-840e-b8eb8757cc13}{Source media}; \href{https://eol.org/media/28315542}{EOL media}; \href{https://eol.org/pages/45510548}{EOL class}; \href{https://creativecommons.org/publicdomain/zero/1.0}{cc-0-1.0}. Display resizing and reconstructed native crop applied; original bytes unchanged. Source-label responses, not expert visibility/species certification.}
\label{fig:rare-same-nonmatching-prediction}
\end{figure}
\clearpage
\begin{figure}[!htbp]\centering
\includegraphics[width=\linewidth]{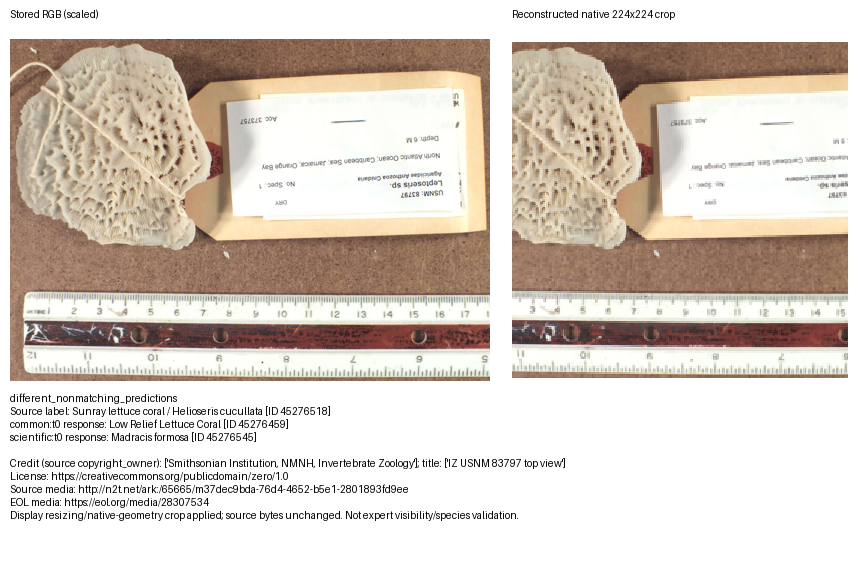}
\caption{different nonmatching predictions. Source label: Sunray lettuce coral / Helioseris cucullata. C-response: Low Relief Lettuce Coral; S-response: Madracis formosa. Credit (publisher field): ['Smithsonian Institution, NMNH, Invertebrate Zoology']. Title: ['IZ USNM 83797 top view']. \href{http://n2t.net/ark:/65665/m37dec9bda-76d4-4652-b5e1-2801893fd9ee}{Source media}; \href{https://eol.org/media/28307534}{EOL media}; \href{https://eol.org/pages/45276518}{EOL class}; \href{https://creativecommons.org/publicdomain/zero/1.0}{cc-0-1.0}. Display resizing and reconstructed native crop applied; original bytes unchanged. Source-label responses, not expert visibility/species certification.}
\label{fig:rare-different-nonmatching-predictions}
\end{figure}

\section{Reproducing the original audits}
\label{app:repro}
The original numerical releases retain image and policy IDs, score matrices, features, class and detector-stratum counts, and development selections. Their SHA records identify the inputs to the tables. The source includes the five credited derivative illustrations. Table~\ref{tab:rare-snapshot} records the eligible Rare rows and exact groups. Each group stays intact in its assigned split.

\begin{table}[!htbp]
\centering
\tablegraphic{\includegraphics{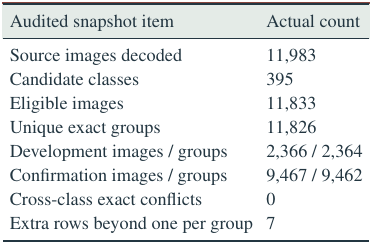}}
\caption{Rare Species image and exact-duplicate group counts.}
\label{tab:rare-snapshot}
\end{table}

The native caches were produced on A100 hardware with FP32, a context length of 77, OpenCLIP 3.2.0 and PyTorch 2.7.0. CPU scoring reuses those image features for the fixed ensemble calculation. Replay recovers the original $t_0$ winners for common, scientific and dual policies by ID. The direct policy checks cover all 308 single-column replacements and the mask diagnostics. The prompt artifact retains the 16 policy predictions and their weighted estimates.

The cached four-template replay covers \CachedReplayFiles{} registered files, \CachedReplayIDs{} predicted-class entries and \CachedReplayChecks{} checks. Deterministic recovery preserves the original predictions, development choices and five illustration IDs. The receipts retain failed versions and their timing. A separate CPU check re-encoded three prompts and one shared development image per checkpoint. Tokens matched, feature and score errors met $2\times10^{-5}$, and all 48 decisions matched against the 154 cached prototypes.

For the external bank, an independent enumeration of within-name pairs reproduced 9,573,336 prediction IDs, 4,453,238 numerical fields, six CSVs and all six development choices. These counts refer to cached bank results. Appendix~\ref{app:v10-repro} gives the distinct checks on fresh image encoding. Two earlier candidate evaluations yielded no species-name outcomes. The CUB study stopped at its geometry-support check \citep{wah2011}, and the iNatLoc study stopped before naming inference.

\end{document}